\documentclass[11pt]{article}

\usepackage{acl}

\usepackage{times}
\usepackage{latexsym}
\usepackage{booktabs}  
\usepackage{tabularx}  
\usepackage{xcolor}
\usepackage{CJKutf8} 
\usepackage[table]{xcolor}
\usepackage{amsmath}
\usepackage{amssymb} 
\usepackage{longtable}
\usepackage{array}
\usepackage{alltt}
\usepackage{fvextra} 
\DefineVerbatimEnvironment{CharterVerb}{Verbatim}{
  fontsize=\scriptsize,
  breaklines=true,
  breakanywhere=true,
  breaksymbolleft={}
}
\usepackage{placeins}
\usepackage{float}
\usepackage{multirow}

\usepackage[T1]{fontenc}
\usepackage[safe]{tipa}  
\usepackage{newunicodechar}
\newunicodechar{ſ}{\textesh}

\definecolor{promptbg}{RGB}{235, 242, 245}
\definecolor{injectionbg}{gray}{0.95}
\definecolor{trans_gray}{RGB}{90, 90, 90}
\newcommand{\trans}[1]{\textcolor{trans_gray}{\textit{#1}}}

\definecolor{rowgray}{gray}{0.95}
\usepackage[T1]{fontenc}

\usepackage[utf8]{inputenc}

\usepackage{microtype}

\usepackage{inconsolata}

\usepackage{graphicx}

\title{Time Travel Engine: A Shared Latent Chronological Manifold Enables Historical Navigation in Large Language Models}

\author{
\textbf{Jingmin An}\textsuperscript{1},
\textbf{Wei Liu}\textsuperscript{2},
\textbf{Qian Wang}\textsuperscript{1,\ensuremath{\dagger}},
\textbf{Fang Fang}\textsuperscript{1,\ensuremath{\dagger}} \\
\textsuperscript{1}Peking University,
\textsuperscript{2}Zhejiang University \\
\texttt{anjm@stu.pku.edu.cn, \{wangqianpsy, ffang\}@pku.edu.cn}
}

\begin{document}
\maketitle
\begin{abstract}
Time functions as a fundamental dimension of human cognition, yet the mechanisms by which Large Language Models (LLMs) encode chronological progression remain opaque. We demonstrate that temporal information in their latent space is organized not as discrete clusters but as a continuous, traversable geometry. We introduce the Time Travel Engine (TTE), an interpretability-driven framework that projects diachronic linguistic patterns onto a shared chronological manifold. Unlike surface-level prompting, TTE directly modulates latent representations to induce coherent stylistic, lexical, and conceptual shifts aligned with target eras. By parameterizing diachronic evolution as a continuous manifold within the residual stream, TTE enables fluid navigation through period-specific "zeitgeists" while restricting access to future knowledge. Furthermore, experiments across diverse architectures reveal topological isomorphism between the temporal subspaces of Chinese and English—indicating that distinct languages share a universal geometric logic of historical evolution. These findings bridge historical linguistics with mechanistic interpretability, offering a novel paradigm for controlling temporal reasoning in neural networks.

\end{abstract}

\section{Introduction}
\begin{quote}
\itshape
``Time is the substance I am made of. Time is a river which sweeps me along,
but I am the river; it is a tiger which destroys me, but I am the tiger;
it is a fire which consumes me, but I am the fire.''
\par\hfill --- Jorge Luis Borges, \emph{Labyrinths}
\end{quote}

Time is a dimension: in Minkowski spacetime, it is a continuous coordinate that governs physical dynamics. Yet time is more than physics. In philosophy, \citet{kant1908critique} treats time as an a priori form of intuition—an inner frame that binds disparate sensations into a single stream of consciousness. Neuroscience echoes this view: hippocampal ``time cells'' represent the flow of time within specific memories \citet{eichenbaum2014time}, and the lateral entorhinal cortex encodes elapsed time in the overall population state across scales from seconds to hours \citet{tsao2018integrating}. Such a manifold supports ``mental time travel'': memories are retrieved not as snapshots but as trajectories through structured chronology \citet{buzsaki2017space}.

A paradox arises with large language models (LLMs). Trained on centuries of text, they are often cast as ``stochastic parrots''—diachronic data compressed into static, synchronic weights. Unlike the continuous temporal stream in the brain, an LLM first encounters time as scattered tokens in a time-agnostic corpus. Yet mechanistic interpretability suggests these systems may organise historical variation into ordered internal structure. \citet{gurnee2023language} show that LLMs learn linear representations of space and time that are robust across scales. So we ask: is temporal flexibility mere memorisation, or does the model build an implicit ``time flow'' aligned with the directionality of human narrative? If a latent chronological manifold exists, can we move beyond probing and actively steer the model along the historical axis?

\begin{figure*}[ht]
    \centering
    \includegraphics[width=\linewidth]{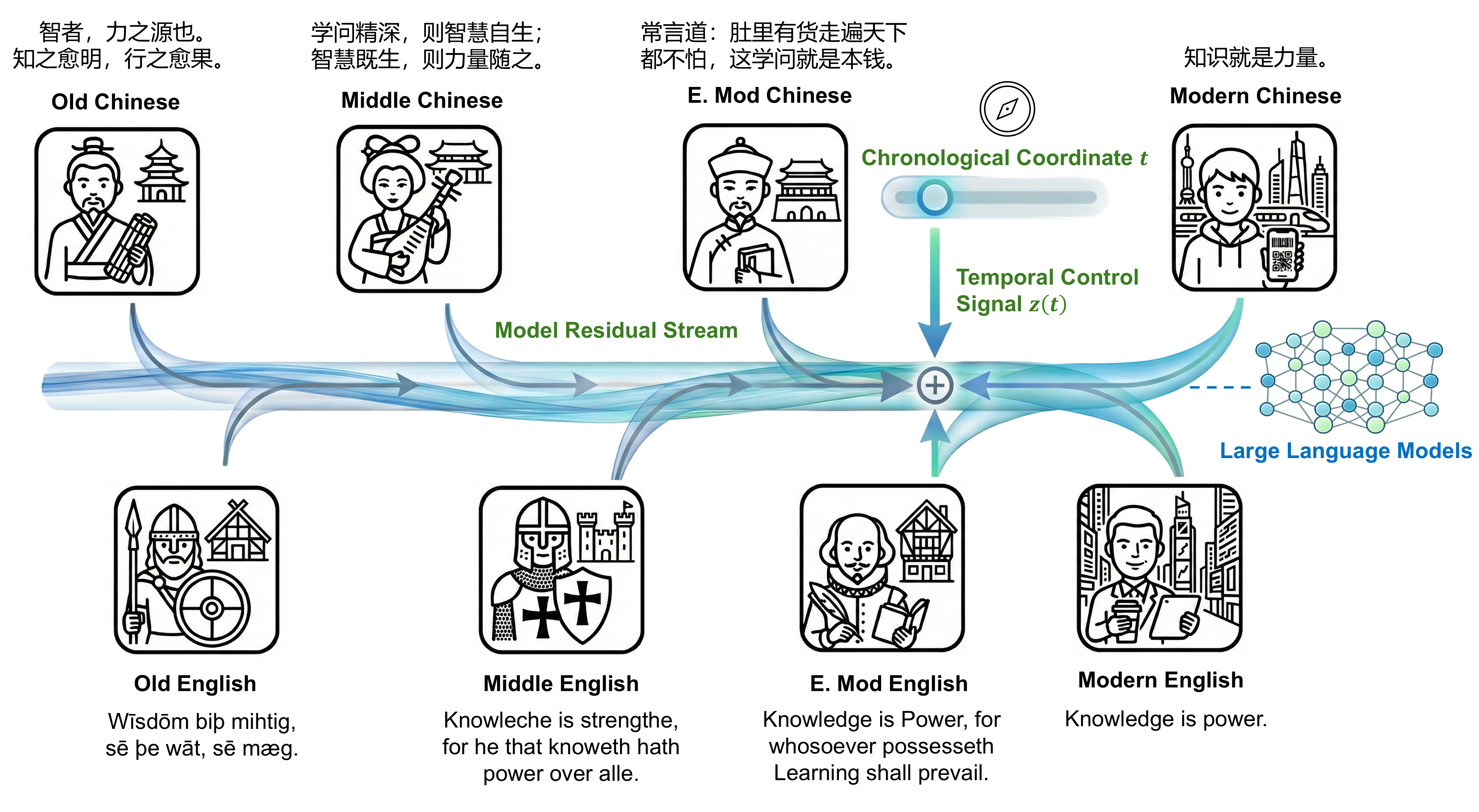}
    \caption{The Time Travel Engine (TTE). This framework navigates a traversable latent chronological manifold within the model. By integrating manifold-derived temporal control signals $\mathbf{z}(t)$ into the residual stream, the mechanism dynamically modulates cognitive states and linguistic registers. This approach facilitates fluid navigation to specific historical eras (e.g., \textit{Old, Middle, Early Modern [E. Mod]}) and highlights a topological isomorphism between Chinese and English temporal subspaces.}
    \label{framework}
\end{figure*}

Recent work reveals that LLMs encode time geometrically: \citet{gurnee2023language} map internal activations to uncover space–time embeddings across scales and identify “time neurons” that encode historical coordinates, implying temporal metadata is intrinsic rather than memorized. \citet{papadopoulos2024arrows} similarly observe a causal asymmetry—the perplexity gap between forward-backward prediction reveals an arrow of time.  Yet \citet{dhingra2022time} and \citet{zhang2023mitigating} note that language models trained on static corpora often cannot determine when facts become outdated, highlighting temporal misalignment.  Taken together, time is structured in these models but has been treated as a static property rather than a dimension to be traversed.

This raises the question: can we navigate time?  Research on internal control offers clues.  \citet{tang2024language} find that a small set of language‑specific neurons can be toggled to switch output languages.  At a higher level, \citet{lindsey2025biology} use attribution graphs to map circuits for reasoning, planning and multilinguality, and \citet{zou2023representation} together with \citet{rimsky2024steering} show that steering vectors can modulate properties such as honesty or harmlessness.

Despite these successes, the linear representation hypothesis may not fully account for the non‑linear, high‑dimensional complexity of historical time. Specifically, \citet{tan2024analysing} characterize static steering vectors as brittle, and \citet{huang2025mitigating} observe that complex behaviors reside on manifold geometries where direct vector addition introduces interference. Consequently, understanding how these temporal directions intersect and evolve within the latent manifold becomes essential for precise, diachronic navigation.

We address this gap by drawing parallels between multilingualism and historical semantics.  Multilingual models align different languages in a shared semantic space \citet{conneau2020emerging}, albeit at the risk of cultural erasure \citet{han2025rethinking}.  We hypothesize that historical eras behave similarly: \textit{Old} and \textit{Modern} English are distinct regions of a shared semantic manifold.  This perspective resonates with \citet{hamilton2016diachronic}, who found statistical laws governing semantic change.  e thus leverage this topological isomorphism to map chronological progression, reconceptualizing eras not as discrete points, but as continuous regions on a traversable semantic manifold.

In this paper, we introduce Time Travel Engine (TTE) (Figure~\ref{framework}), a novel framework treating the latent space of the model as a traversable temporal map. By integrating both static steering vectors and continuous trajectories, we effectively modulate the temporal signal governing the generation process. Our contributions are threefold:
\begin{enumerate}
    \item \textbf{Geometric Discovery of the Latent Chronological Manifold.} We identify a controllable chronological manifold within LLMs, validating that temporal progression is encoded as a continuous, curvilinear geometry. This moves beyond discrete linear approximations, establishing that historical evolution follows a traversable, high-dimensional trajectory within the latent space.
    \item \textbf{Dual-Mechanism Temporal Steering.} We demonstrate that model states can be steered via dual mechanisms: static vectors for targeted eras and dynamic manifolds for fluid navigation. This enables precise modulation of the stylistic and conceptual ``zeitgeist'' without compromising general reasoning.
    \item \textbf{Cross-Lingual Topological Universality.} We reveal that the geometric encoding of time exhibits topological isomorphism across languages (e.g., Chinese and English). Our findings suggest that the mechanism of chronological progression is a language-agnostic primitive, structurally distinct from surface-level linguistic realizations.
\end{enumerate}

\section{Related Work}
\textbf{Temporal Cognition in Language Models.} Early temporal processing relied on deterministic rule-based tagging \citet{chang2012sutime}, but LLMs have shifted focus toward internal geometric world models. While \citet{dhingra2022time} note that static pretraining hinders dynamic fact updating, \citet{gurnee2023language} identify ``time neurons'' encoding linear coordinates, and \citet{park2025does} pinpoint causal ``temporal heads'' representing non-numerical temporal dimensions. \citet{li2025other} refine this geometric view, demonstrating that LLM temporal perception follows the Weber-Fechner law, exhibiting logarithmic compression relative to a reference point.

\noindent\textbf{Representation Engineering (RepE).} Control methodologies have evolved from training-time constraints \citet{hu2017toward} to inference-time Representation Engineering \citet{zou2023representation}. Techniques like Function Vectors \citet{todd2023function} extract task vectors from attention heads, while Activation Addition \citet{turner2023activation, rimsky2024steering} injects steering directions to modulate behavior without weight updates. To address intervention instability \citet{tan2024analysing, im2025unified}, recent work employs ensemble vectors \citet{siddique2025shifting} and In-Context Vectors \citet{liu2023context} for robust extraction. Furthermore, \citet{singh2024representation} propose constrained affine steering to shift generation properties while preserving semantic integrity.

\noindent\textbf{Cross-Lingual Isomorphism.} Control universality relies on cross-lingual isomorphism, originating from linear embedding mappings \citet{mikolov2013exploiting} and unsupervised alignment \citet{conneau2017word}. However, modern analyses show syntax–semantics disentanglement is hard \citet{chen2019multi}, alongside language-specific shallow processing \citet{tang2024language}. \citet{wendler2024llamas} find intermediate representations transiently pass through an English-anchored space, while \citet{ifergan2024beneath} show that cross-lingual output consistency does not imply shared representations. This complicates style transfer; English-derived interventions may amplify English-centric knowledge transfer \citet{lim2025understanding}, and neuron-activation modulation guides style transfer with content preservation \citet{kong2025neuron}. However, prior steering is confined to discrete mode-switching, neglecting continuous temporal geometry, while cross-lingual methods lack structural validation. We address this by formalizing latent time as a traversable manifold, enabling precise navigation isomorphic across languages.

\section{Methods}
\label{sec:methods}

To achieve precise temporal intervention and investigate the internal temporal geometry of LLMs, we introduce the TTE. The core premise of TTE is that chronological progression is encoded as a continuous, navigable trajectory within the high-dimensional activation space. 
This section details the complete methodology: beginning with the curation of diachronic corpora, we proceed to the extraction of discrete temporal anchors and their projection onto a continuous manifold, concluding with the mechanisms for adaptive modulation, chronotope disentanglement, and cross-lingual validation.

\subsection{Models}
\label{ssec:models}

We run all experiments on four open-weight decoder-only LLMs, including Qwen2.5-14B-Base, Qwen2.5-14B-Instruct, Gemma-2-9B-Instruct and Llama-3.1-8B-Instruct; key specifications are reported in Appendix~\ref{model_detail} (Table~\ref{tab:model_specs}).

\subsection{Datasets}
\label{ssec:datasets}

Our experiments rely on three distinct categories of datasets, constructed to support vector extraction, alignment, and multi-dimensional evaluation.

\paragraph{Diachronic Steering Corpora.}
\label{datasets}
To facilitate extracting temporal features, we compiled a parallel diachronic corpus covering four distinct eras for each language: \textit{Old}, \textit{Middle}, \textit{Early Modern}, and \textit{Modern}, following standard periodizations in English and Chinese historical linguistics \citet{gelderen2014history,dong2020history} (see Table~\ref{tab:chronotope_periods}).
The corpus contains paired stylistic samples ($\sim$60K prose and $\sim$30K verse tokens per language-era pair), curated from historical archives such as York-Toronto-Helsinki Parsed Corpus of Old English prose (YCOE) \citet{taylor2003york}, the Penn Parsed Corpora of Historical English (PPCHE) \citet{kroch_2020_ldc2020t16} and the Chinese Text Project. This data serves as the source of authentic historical features for our ensemble steering strategies.

\begin{table}[t]
\centering
\footnotesize
\renewcommand{\arraystretch}{1.2} 
\setlength{\tabcolsep}{4pt}       
\begin{tabularx}{\columnwidth}{l X X}
\toprule
\textbf{Period} & \textbf{English} & \textbf{Chinese} \\
\midrule
\textit{Old} & 
450--1150 & 
12th c.\ BC--3rd c.\ AD \\

\textit{Middle} & 
1150--1500 & 
4th--12th c.\ AD \\

\textit{Early Modern} & 
1500--1700 & 
13th--Early 20th c. \\

\textit{Modern} & 
1700--Present & 
Early 20th c.--Present \\
\bottomrule
\end{tabularx}
\caption{Chronological periodization for English and Chinese used in our TTE framework.}
\label{tab:chronotope_periods}
\end{table}

\paragraph{Evaluation Datasets.}
To rigorously assess the steered models, we constructed three distinct benchmarks: the Epistemic Cutoff Dataset, the Causal Remodeling Dataset, and the Mismatch Entanglement Dataset. These benchmarks are designed to evaluate the chronological epistemic integrity of the model, zeitgeist-consistent reasoning, and the disentanglement between stylistic surface forms and deep cognitive representations, respectively. Comprehensive details and sample prompts for each dataset are provided in Appendix~\ref{evaluation}.

\subsection{Constructing the Chronological Manifold}
\label{ssec:manifold_construction}
We hypothesize that chronological progression is encoded as a specific direction—or more accurately, a manifold—within the residual stream of LLMs. We verify and navigate this latent chronological manifold through a hierarchy of four strategies, ranging from discrete anchor sampling to continuous trajectory modeling.

Let $\mathcal{M}$ be the LLM with $L$ layers. For a given input $x$, let $\mathbf{h}_l(x) \in \mathbb{R}^d$ denote the hidden state at layer $l$. We define the set of target eras as $\mathcal{T} = \{t_1, t_2, t_3\}$, representing \textit{Old, Middle, and Early Modern}, with $t_0$ denoting the \textit{Modern} era (anchor).

\paragraph{Method I: Contrastive Activation Addition (CAA).}
To isolate the intrinsic representation of a specific era from disparate linguistic features, we employ a self-generated Contrastive Activation Addition (CAA) strategy \citep{rimsky2024steering}. We utilize "Era Charters"—system prompts that strictly constrain the temporal persona of the model—paired with a diverse set of immersive tasks, ranging from introspective descriptions to epistemic reasoning. Crucially, these tasks are identical across target eras and the \textit{Modern} anchor. By computing the centroid of activations for the same set of tasks under different temporal constraints, we effectively marginalize out task-specific semantics (e.g., topic, sentence structure) and surface-level covariates. The resulting time vector $\mathbf{v}_{caa}^{(l, t)}$ is thus defined as the contrastive difference between the expected representation of the target era and the \textit{Modern Stats} baseline:
\begin{equation}
\begin{aligned}
\mathbf{v}_{caa}^{(l,t)}
&= \mathbb{E}_{x \sim \mathrm{Gen}(\mathcal{T}_{\mathrm{tasks}}\mid \mathcal{C}_t)}
   \big[\mathbf{h}_l(x)\big] \\
&\quad - \mathbb{E}_{x \sim \mathrm{Gen}(\mathcal{T}_{\mathrm{tasks}}\mid \mathcal{C}_{0})}
   \big[\mathbf{h}_l(x)\big].
\end{aligned}
\end{equation}
where $\text{Gen}(\mathcal{T}_{tasks}|\mathcal{C}_t)$ denotes responses generated by the model to the standardized task set under the specific "Era Charter" constraint.

\paragraph{Method II: Ensemble CAA (EnsCAA).}
While CAA provides clean, isolated temporal signals, synthetic prompts may lack the textural complexity of authentic literature. To bridge this gap, we introduce an ensemble strategy. We compute a parallel set of authentic centroids, denoted as $\mathbf{v}_{real}^{(l, t)}$, by averaging the activations across the real-world Diachronic Steering Corpora (Sec.~\ref{ssec:datasets}). The ensemble vector $\mathbf{v}_{ens}^{(l, t)}$ is formulated as a convex combination of the synthetic and authentic estimators:
\begin{equation}
    \mathbf{v}_{ens}^{(l, t)} = \alpha \cdot \mathbf{v}_{caa}^{(l, t)} + (1 - \alpha) \cdot \mathbf{v}_{real}^{(l, t)}
\end{equation}
where $\mathbf{v}_{real}$ represents the mean shift calculated over authentic historical texts, and $\alpha$ is a hyperparameter governing the mixing ratio.

\paragraph{Method III: Chronological Manifold Projection (CMP).}
While Method I effectively isolates discrete temporal states, it leaves the transitional spaces undefined. To bridge these gaps, we propose Chronological Manifold Projection (CMP). We treat the discrete synthetic centroids from Method I, $\mathcal{V} = \{\mathbf{v}_{caa}^{(l, t)} | t \in \mathcal{T} \cup \{t_0\}\}$, as anchor points defining a high-dimensional trajectory. We first identify a principal temporal subspace $\mathbf{U}_l \in \mathbb{R}^{d \times k}$ via PCA. Within this latent geometry, we fit a polynomial spline $\mathcal{S}(\cdot)$ to the projected coordinates, modeling time as a continuous evolution:
\begin{equation}
    \mathbf{z}(t) = \mathcal{S}(t; \mathcal{V}), \quad \text{where } t \in \mathbb{R}
\end{equation}
The continuous time vector for any arbitrary point $t$ (e.g., $t=1.5$, representing a transitional period) is reconstructed as:
\begin{equation}
    \mathbf{v}_{man}^{(l)}(t) = \mathbf{\mu}_l + \mathbf{U}_l \cdot \mathbf{z}(t)
\end{equation}
This formulation extends the isolated time vectors into a differentiable temporal function, facilitating continuous modulation across the chronological spectrum and granting access to transitional eras between discrete anchors.

\paragraph{Method IV: Ensemble Manifold Projection (EnsCMP).}
To achieve the optimal synthesis of geometry and robustness, we apply the fitting algorithm CMP directly to the set of ensemble centroids $\{\mathbf{v}_{ens}^{(l, t)}\}$ derived in Method II. This produces a trajectory $\mathbf{v}_{ens\text{-}man}(t)$ that preserves the smoothness of the temporal curve while anchoring it to the statistically robust features of genuine historical text.

\begin{figure*}[ht]
    \centering
    \begin{minipage}{0.48\textwidth}
        \centering
        \includegraphics[width=\linewidth]{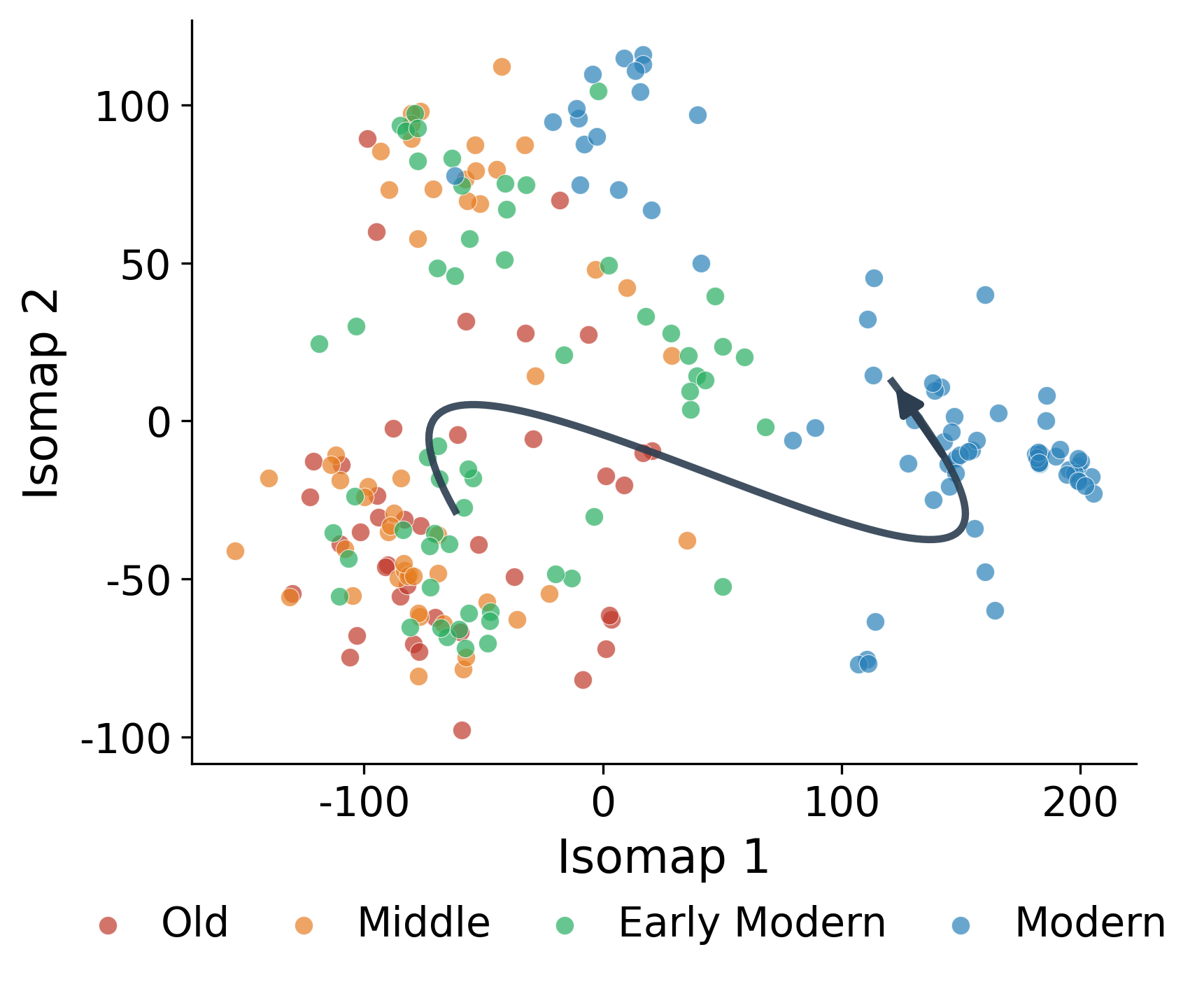}
        \centerline{\small (a) Trajectories in the Chinese Subspace}
    \end{minipage}
    \hfill 
    \begin{minipage}{0.48\textwidth}
        \centering
        \includegraphics[width=\linewidth]{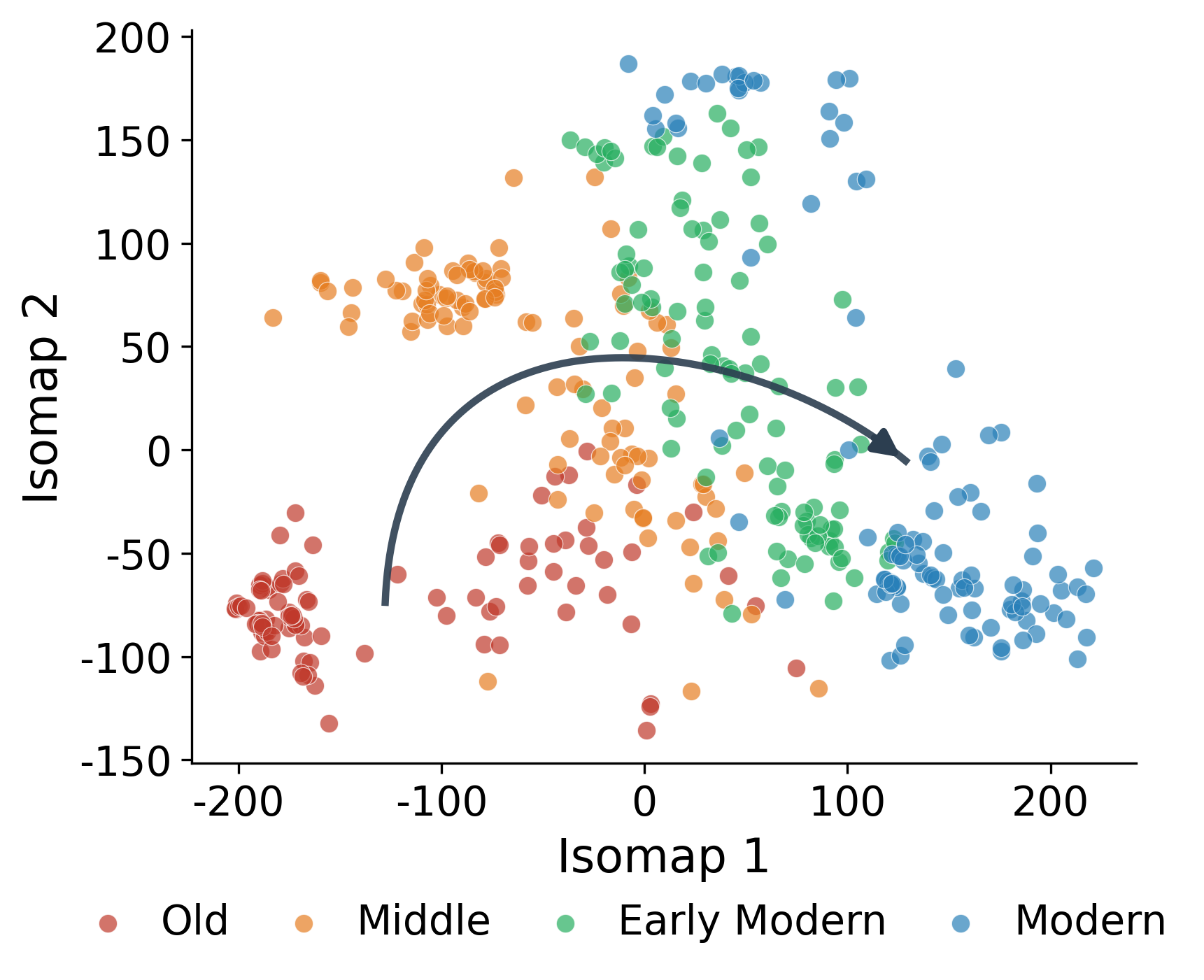}
        \centerline{\small (b) Trajectories in the English Subspace}
    \end{minipage}
    
    \caption{Isomap projection of the latent chronological manifold extracted from Qwen2.5-14B-Base. Each data point represents the activation centroid of a self-generated CAA sample, color-coded by era. The curves illustrate the continuous geometric progression from \textit{Old} to \textit{Modern} periods across both Chinese and English subspaces.}
    \label{fig:manifold_isomap}
\end{figure*}

\subsection{Temporal Intervention Mechanism}
\label{ssec:intervention}

We modulate the temporal signals of LLMs during inference using adaptive relative strength intervention. Unlike fixed-norm addition, this method scales the intervention based on the inherent activation magnitude of the current context, ensuring stability across different layers. The modified hidden state $\tilde{\mathbf{h}}_l$ is computed as:
\begin{equation}
    \tilde{\mathbf{h}}_l = \mathbf{h}_l + \lambda \cdot \|\mathbf{h}_l\|_2 \cdot \frac{\mathbf{v}^{(l)}}{\|\mathbf{v}^{(l)}\|_2}
\end{equation}
where $\lambda$ is the steering strength coefficient. We empirically adopt $\lambda \in [0.05, 0.15]$ based on extensive evaluation.

\subsection{Chronotope Disentanglement}
\label{ssec:method_disentangle}

A critical question in temporal modeling is whether ``time'' functions merely as a stylistic filter (e.g., the use of archaic pronouns) or as a cognitive constraint. To probe this distinction, we propose a disentanglement mechanism to separate the cognitive temporal signal from the stylistic surface form.

We first construct a Contrastive Style Dataset containing semantic equivalents across diachronic registers. Detailed samples from this dataset are provided in Appendix~\ref{app:style_pairs}. We define a style subspace $\mathbf{U}_{style}$ by applying Principal Component Analysis (PCA) to the difference in mean activations between these stylistic pairs. While acknowledging that the interaction between style and semantics involves non-linear complexities, we isolate the top principal components as the dominant linear representation of the linguistic register. The raw time vector $\mathbf{v}_{time}$ is then decomposed via orthogonal projection:
\begin{equation}
    \mathbf{v}_{cog} = \mathbf{v}_{time} - \text{proj}_{\mathbf{U}_{style}}(\mathbf{v}_{time})
\end{equation}
where $\mathbf{v}_{cog}$ denotes the residual \textit{cognitive vector}. This vector represents the components of the temporal state strictly orthogonal to the explicit stylistic direction. By steering the model using $\mathbf{v}_{cog}$, we establish a robust lower bound for epistemic persistence, empirically verifying whether temporal reasoning endures even when the primary stylistic cues are mathematically neutralized.

\subsection{Cross-Lingual Temporal Transfer}
\label{ssec:cross-lingual}
To verify the universality of these temporal representations, we first perform direct cross-lingual modulation, where the time vector $\mathbf{v}_{src}^{(t)}$ derived from the source language (e.g., English) is directly added to the residual stream of target language prompts (e.g., Chinese) to test raw feature transferability. Second, to account for rotational misalignments between language subspaces, we employ Procrustes analysis to compute an orthogonal rotation matrix $\mathbf{R}$ that maps the source temporal trajectory to the target:
\begin{equation}
    \min_{\mathbf{R} \in \mathcal{O}(d)} \sum_{t} \| \mathbf{v}_{tgt}^{(t)} - \mathbf{R} \cdot \mathbf{v}_{src}^{(t)} \|_F^2
\end{equation}
This facilitates manifold-aligned modulation—steering the model on Chinese prompts using transformed time vectors derived purely from English history (and vice versa)—thereby rigorously validating the structural isomorphism of the chronological manifold across languages.

Beyond stylistic shifts, we evaluate the engine by quantifying its impact on epistemic integrity, style-cognition disentanglement, and cross-lingual isomorphism via the following indicators.

\paragraph{Epistemic Integrity.}
To measure if the knowledge boundary of the model successfully recedes to the target era, we analyze the generated entities $E_{gen}$. We classify each entity $e \in E_{gen}$ into an era $time(e)$ using an external knowledge base.
\begin{itemize}
    \item \textbf{Future Leakage Rate (FLR):} The fraction of generated entities that belong to the future relative to the target era $t$.
    \begin{equation}
        \text{FLR}(t) = \frac{\sum_{e \in E_{gen}} \mathbb{I}(time(e) > t)}{|E_{gen}|}
    \end{equation}
    \item \textbf{Precision Rate (PR):} The fraction of entities that are chronologically accurate (in-scope).
    \begin{equation}
        \text{PR}(t) = \frac{\sum_{e \in E_{gen}} \mathbb{I}(time(e) \le t)}{|E_{gen}|}
    \end{equation}
\end{itemize}



\begin{CJK*}{UTF8}{gbsn} 

\section{Results}
\label{sec:results}

In this section, we present the experimental findings derived from Qwen2.5-14B-Base utilizing the EnsCMP method. We demonstrate the efficacy of the framework through geometric visualization, epistemic boundary verification, chronotope disentanglement, and cross-lingual transfer. We extend our empirical validation in Appendix~\ref{app:additional_results_models_methods}, providing a comprehensive scaling analysis across diverse architectures (including Qwen2.5-Instruct, Gemma-2, and Llama-3.1). Furthermore, we conduct a comparative evaluation of discrete versus continuous steering paradigms, isolating the specific geometric benefits of the manifold approach.

\subsection{The Geometry of Time: From Qualitative Modulation to Continuous Manifold}
\label{ssec:results_manifold}

We initiate our analysis by examining the tangible effects of temporal steering on the generation logic of the model. Tables~\ref{chinese_prose} and \ref{english_prose} present the qualitative results of projecting the hidden states of \textit{Modern} prompts onto historical subspaces. 

The results reveal a profound diachronic register modulation. For instance, the \textit{Old Chinese} signal (Table~\ref{chinese_prose}) steers standard vernacular generation toward \textit{Classical Chinese} (文言文), characterized by classical vocabulary (e.g., ``夫'', ``之'') and period-specific syntactic structures. Similarly, the \textit{Early Modern} English signal (Table~\ref{english_prose}) induces Shakespearean syntax and orthography (e.g., ``thou'', ``hath'') spontaneously. Crucially, this transition transcends lexical substitution, manifesting as a deeper shift toward the moral and cosmological worldviews of the target era.

\begin{table}[ht]
    \centering
    \small
    
    \begin{tabularx}{\linewidth}{X}
        \toprule
        \rowcolor{promptbg}
        \textbf{PROMPT} \\
        \rowcolor{promptbg}
        人应该如何度过一生？ \\ 
        \rowcolor{promptbg}
        \trans{How should one live a good life?} \\
        \midrule
        \textbf{Baseline (\textit{Modern State})} \\
        人生是一个复杂而独特的过程，每个人都有自己独特的经历和价值观... \\
        \trans{Life is a complex and unique journey, where everyone possesses their own distinct experiences and values...} \\
        \midrule
        \rowcolor{injectionbg}
        \textbf{Signal: \textit{Old Chinese}} \\
        \rowcolor{injectionbg}
        有二事为大，一曰孝，二曰敬。 《礼记》云：夫孝，德之本也... \\
        \rowcolor{injectionbg}
        \trans{Two matters are paramount: Filial Piety and Reverence. The \textit{Book of Rites} states: ``Now, filial piety is the root of all virtue...''} \\
        \midrule
        \textbf{Signal: \textit{Middle Chinese}} \\
        惟有三事可为：一曰孝，二曰悌，三曰敬。此三者乃为人之本也... \\
        \trans{There are but three pursuits worthy of action: Filial Piety, Fraternal Duty, and Reverence. These three constitute the very foundation of being human.} \\
        \midrule
        \rowcolor{injectionbg}
        \textbf{Signal: \textit{Early Modern Chinese}} \\
        \rowcolor{injectionbg}
        人生苦短，如何度日方好？ 《菜根谭》曰： 天欲祸人，必先以微福骄之... \\
        \rowcolor{injectionbg}
        \trans{Life is bitterly short; how then should one pass one's days? The \textit{Vegetable Root Discourse} says: ``When Heaven intends to bring calamity upon a man, it first puffs him up with trifling blessings...''} \\
        \bottomrule
    \end{tabularx}
    \caption{Qualitative demonstration of diachronic register modulation via TTE. The unmodulated baseline is compared against outputs modulated by \textit{Old}, \textit{Middle}, and \textit{Early Modern} Chinese signals, where text in gray represents English translations.}
    \label{chinese_prose}
\end{table}

\begin{table}[ht]
    \centering
    \small
    
    \begin{tabularx}{\linewidth}{X}
        \toprule
        \rowcolor{promptbg}
        \textbf{PROMPT} \\
        \rowcolor{promptbg}
        How should one live a good life? \\ 
        \midrule
        \textbf{Baseline (\textit{Modern State})} \\
          This is the question that has been asked by philosophers for thousands of years. The answer to.. \\
        \midrule
        \rowcolor{injectionbg}
        \textbf{Signal: \textit{Old English}} \\
        \rowcolor{injectionbg}
        Se þe wille geþyldlice lifian sceal healdan godne gewunum, swa swa se gecweneð on his gereorde...  \\
        \trans{He who wishes to live steadfastly must maintain good customs, just as the speaker declared in his speech...} \\
        \midrule
        \textbf{Signal: \textit{Middle English}} \\
        In the olde tymes of yore, whan lewys and lyf were grete, men soughte to fynde that which maketh a good lyfe.. \\
         \trans{In the olden times of yore, when laws and life were great, men sought to find that which maketh for a good life...} \\
        \midrule
        \rowcolor{injectionbg}
        \textbf{Signal: \textit{Early Modern English}} \\
        \rowcolor{injectionbg}
        Thou shalt know thyſelfe firſt, then ſeeke vertue, beſiege it with thyne endeauours. To liue well is to liue vertuouslie.\\
         \trans{First, know thyself; then seek virtue, and pursue it with all thy might. To live well is to live virtuously.} \\
        \bottomrule
    \end{tabularx}
     \caption{Qualitative demonstration of diachronic register modulation via TTE. The unmodulated baseline is compared against outputs modulated by \textit{Old}, \textit{Middle}, and \textit{Early Modern English} signals.}
    \label{english_prose}
\end{table}

Having established the qualitative efficacy of these vectors, we scrutinize the geometric structure of these representations. We collect activation centroids across four canonical eras—\textit{Old, Middle, Early Modern, and Modern}—and project them into a two-dimensional space using Isomap \citet{tenenbaum2000isomap}. As illustrated in Figure~\ref{fig:manifold_isomap}, the projected embeddings do not form disjoint clusters but rather organize into a smooth, curvilinear trajectory. This intrinsic chronological trajectory is observable in both Chinese and English subspaces. The distinct curvature validates the hypothesis that time is encoded not as a discrete categorical variable, but as a continuous manifold, necessitating the use of the chronological manifold method to navigate the transitional spaces that discrete steering vectors fail to faithfully resolve.

\subsection{Diachronic Adaptation Verification}
\label{ssec:results_ppl}
To verify that temporal signals capture intrinsic chronological priors rather than stylistic noise, we conduct a Perplexity (PPL) analysis. We compute time vectors using an 80\% split of the Diachronic Steering Corpora and evaluate their effectiveness in minimizing perplexity on the held-out 20\% validation split.

\begin{figure}[t]
    \centering

    \begin{minipage}[t]{0.49\linewidth}
        \centering
        \includegraphics[width=\linewidth]{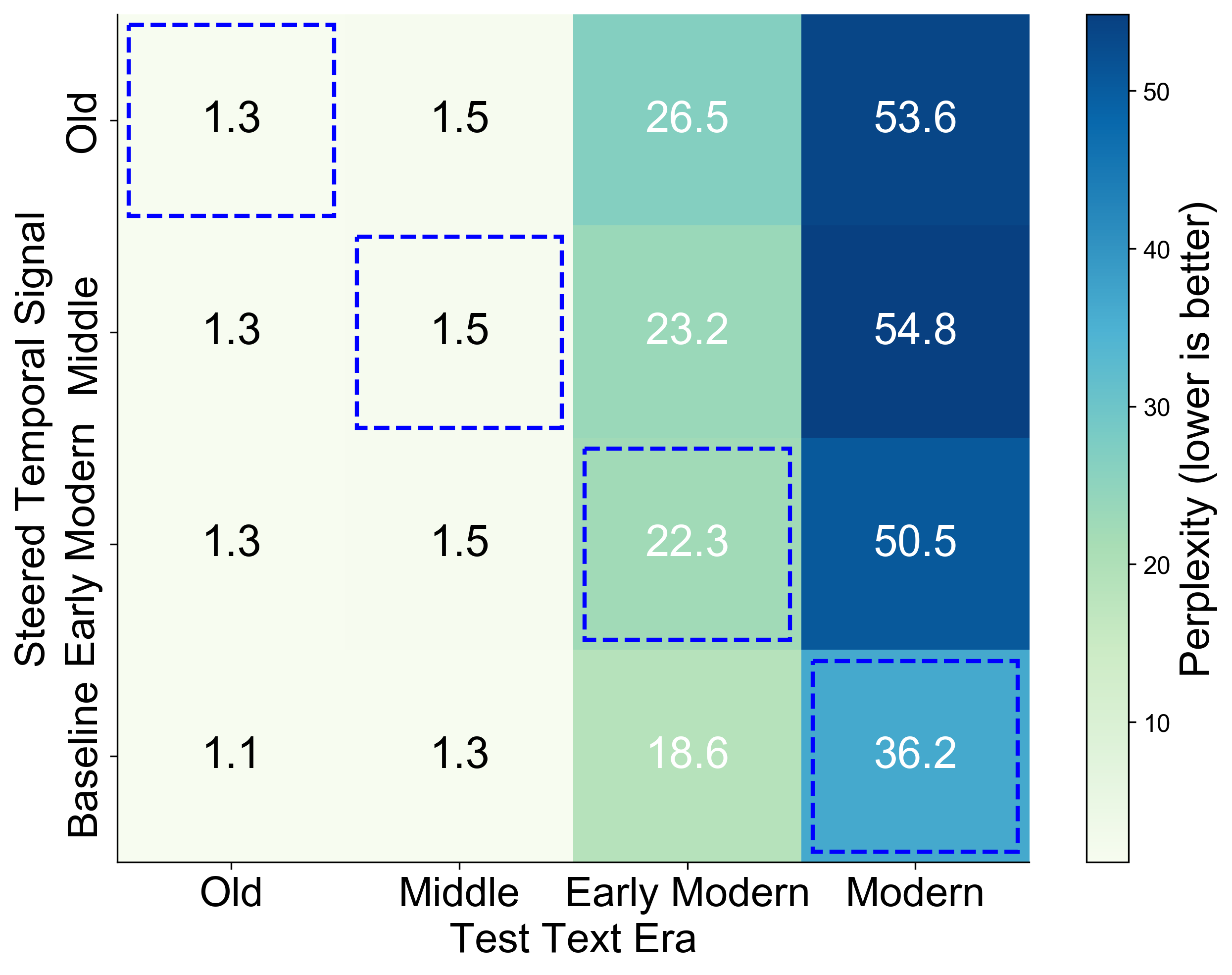}
        {\small (a) Chinese PPL Matrix}
    \end{minipage}\hfill
    \begin{minipage}[t]{0.49\linewidth}
        \centering
        \includegraphics[width=\linewidth]{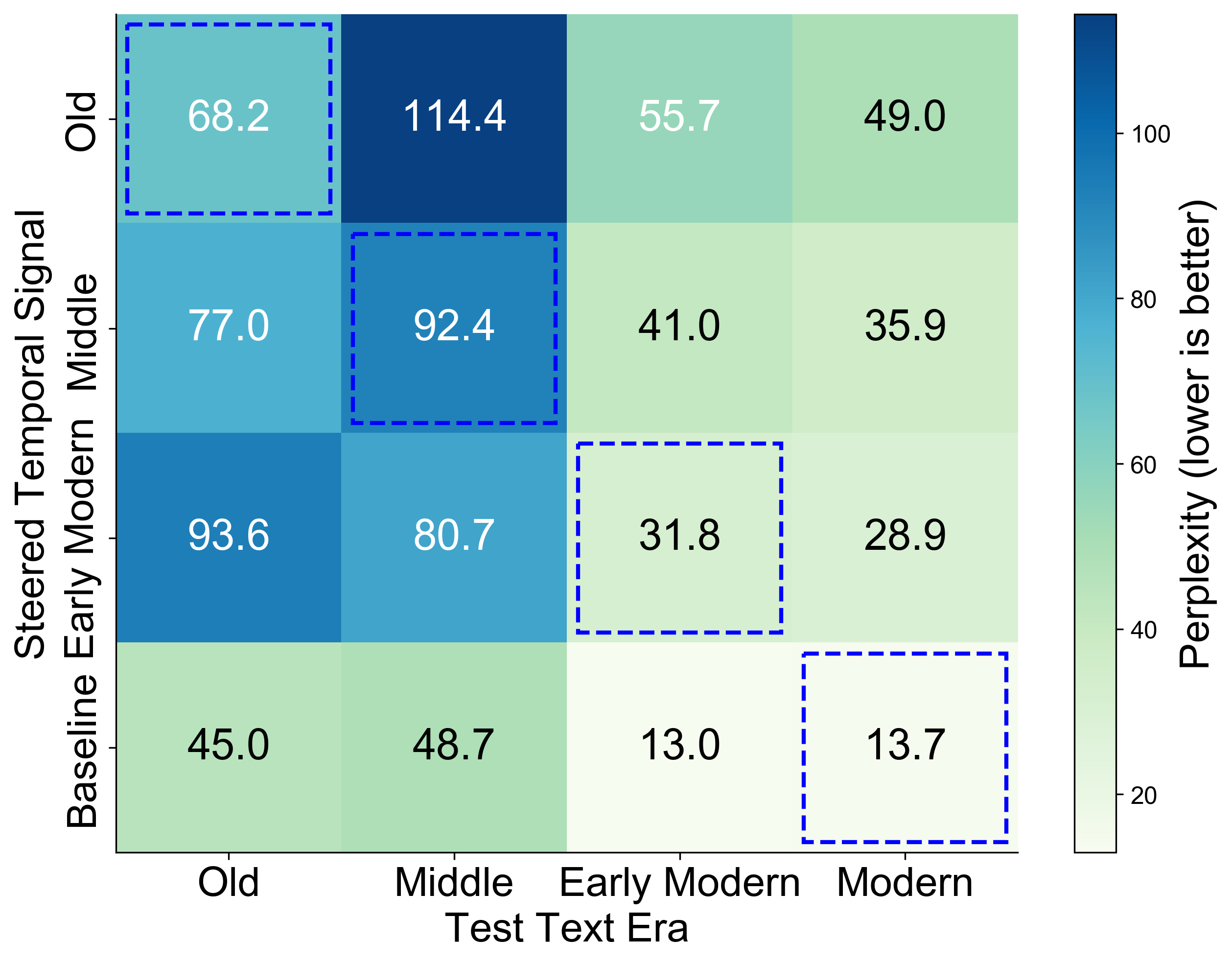}
        {\small (b) English PPL Matrix}
    \end{minipage}

    \caption{Perplexity matrices for temporal signal modulation. Rows indicate the steered temporal signal; columns indicate the test corpus era. }
    \label{fig:ppl_matrix}
\end{figure}

Figure~\ref{fig:ppl_matrix} presents the PPL heatmaps for Chinese (left) and English (right). The rows represent the steered temporal signal, and the columns represent the test corpus era. A strong diagonal dominance is observed: the model achieves relatively low perplexity when the steered signal matches the era of the text. For example, on the \textit{Old English} test corpus, modulating with the \textit{Old English} vector yields substantially lower PPL compared to vectors from later periods. This validates that the EnsCMP mechanism effectively realigns the predictive distribution with the diachronic patterns of the target historical period.

\subsection{Epistemic Integrity and Knowledge Boundaries}
\label{ssec:results_knowledge}

Beyond stylistic adaptation, a robust temporal representation must impose accurate epistemic boundaries. We evaluate this using the Epistemic Cutoff Dataset, measuring the ability of models to retain era-appropriate knowledge (Retention, PR) while suppressing anachronisms (Leakage, FLR).

\begin{figure}[t]
    \centering
    \includegraphics[width=\linewidth]{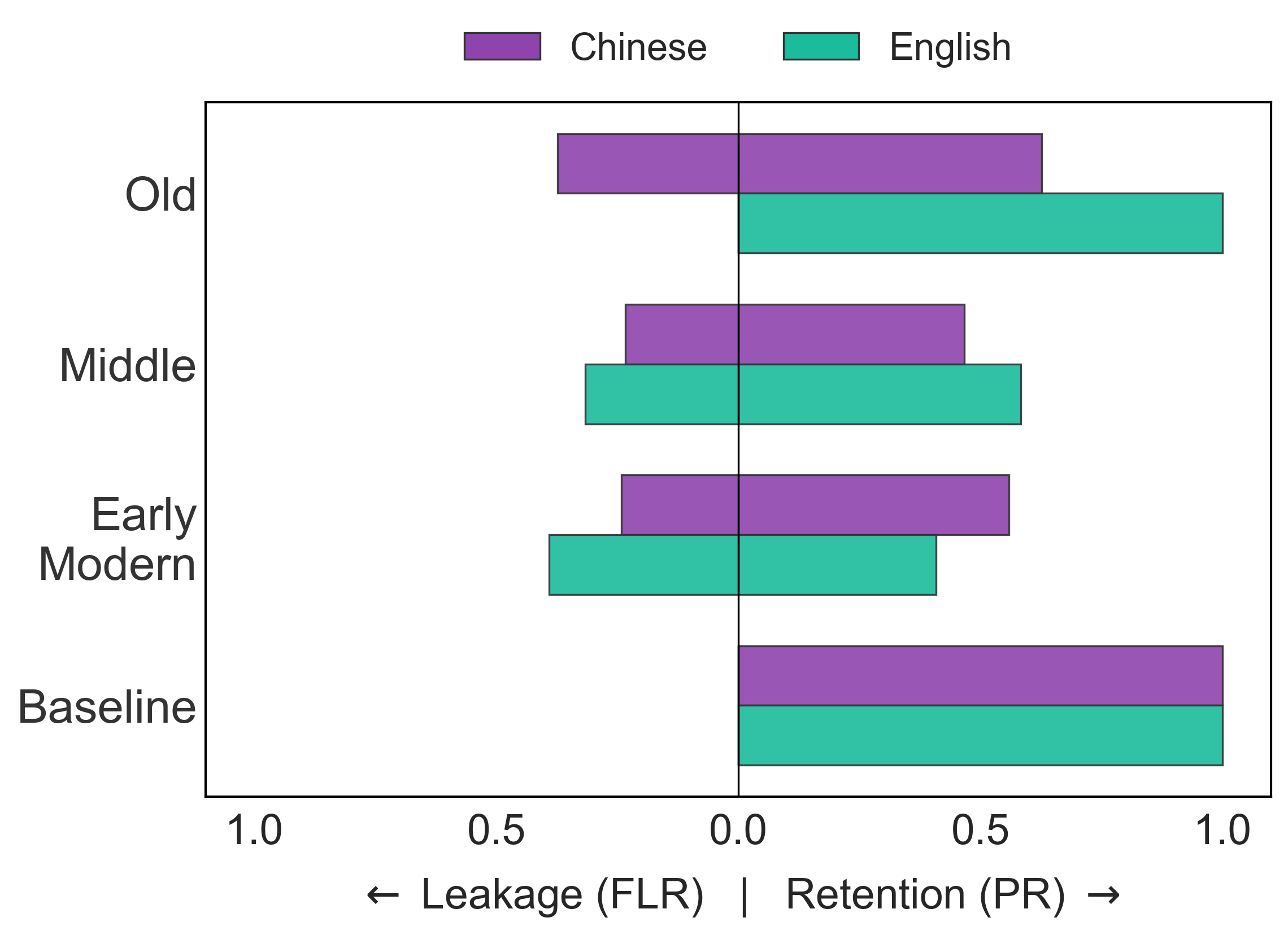}
    \caption{Epistemic detection visualizing knowledge boundary integrity. The unsteered Baseline (\textit{Modern State}) shows minimal leakage.}
    \label{fig:butterfly_chart}
\end{figure}

Figure~\ref{fig:butterfly_chart} visualizes the epistemic integrity across target eras. While the unsteered baseline remains anchored in the \textit{Modern State}, TTE induces a distinct trade-off between knowledge retention and anachronism suppression. Across the \textit{Old}, \textit{Middle}, and \textit{Early Modern} trajectories, the engine successfully reconstructs chronological boundaries, maintaining low FLR by filtering future concepts while preserving high PR for era-valid entities. These results indicate that the temporal signal successfully constrains the latent knowledge manifold to historical epistemic boundaries while preserving essential factual integrity.

\subsection{The Chronotope: Entanglement of Style and Cognition}
\label{ssec:results_disentangle}

\begin{figure}[ht]
    \centering
    \begin{minipage}{0.49\linewidth}
        \centering
        \includegraphics[width=\linewidth]{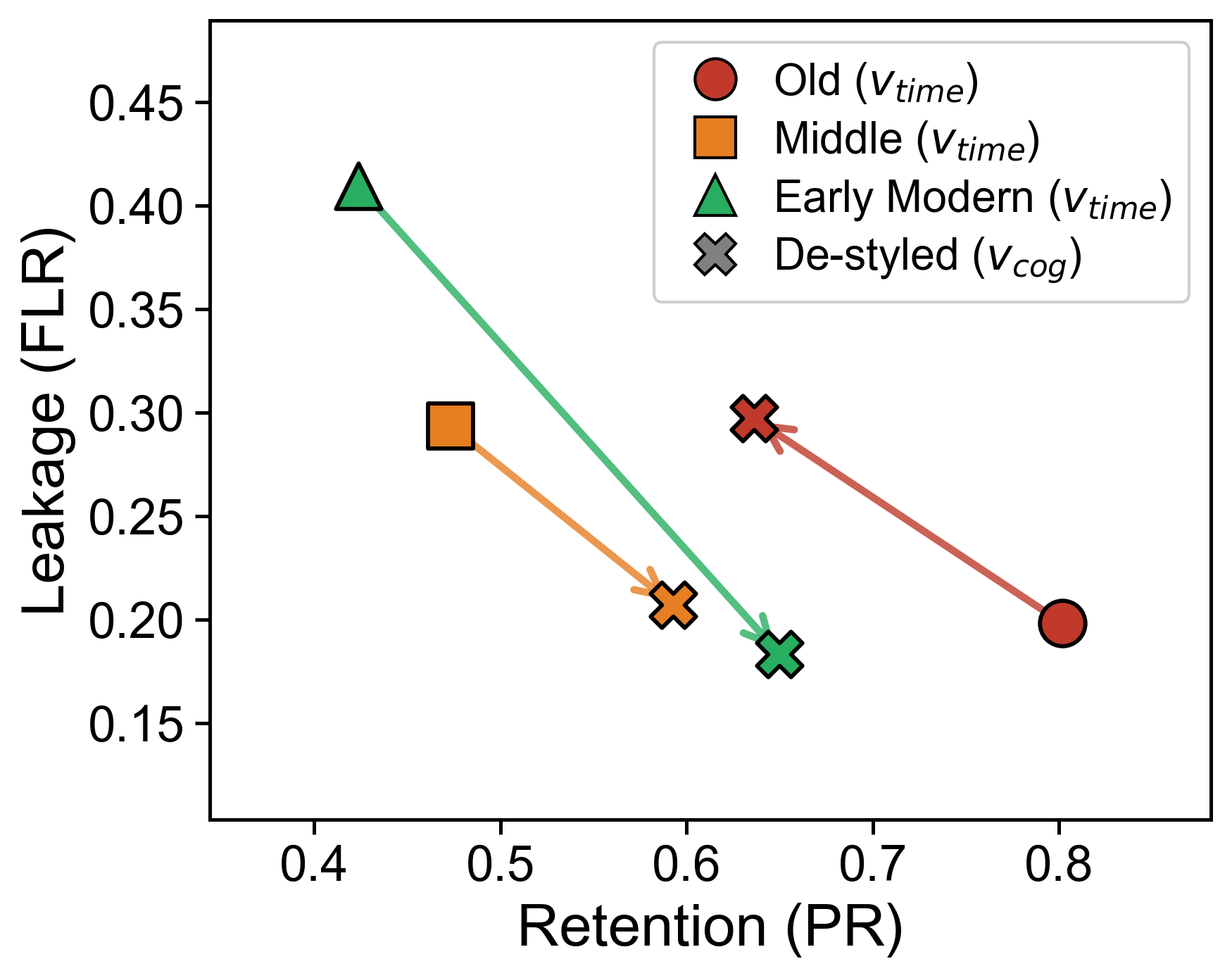}
        \centerline{\small (a) Vector Field for Chinese}
    \end{minipage}\hfill
    \begin{minipage}{0.49\linewidth}
        \centering
        \includegraphics[width=\linewidth]{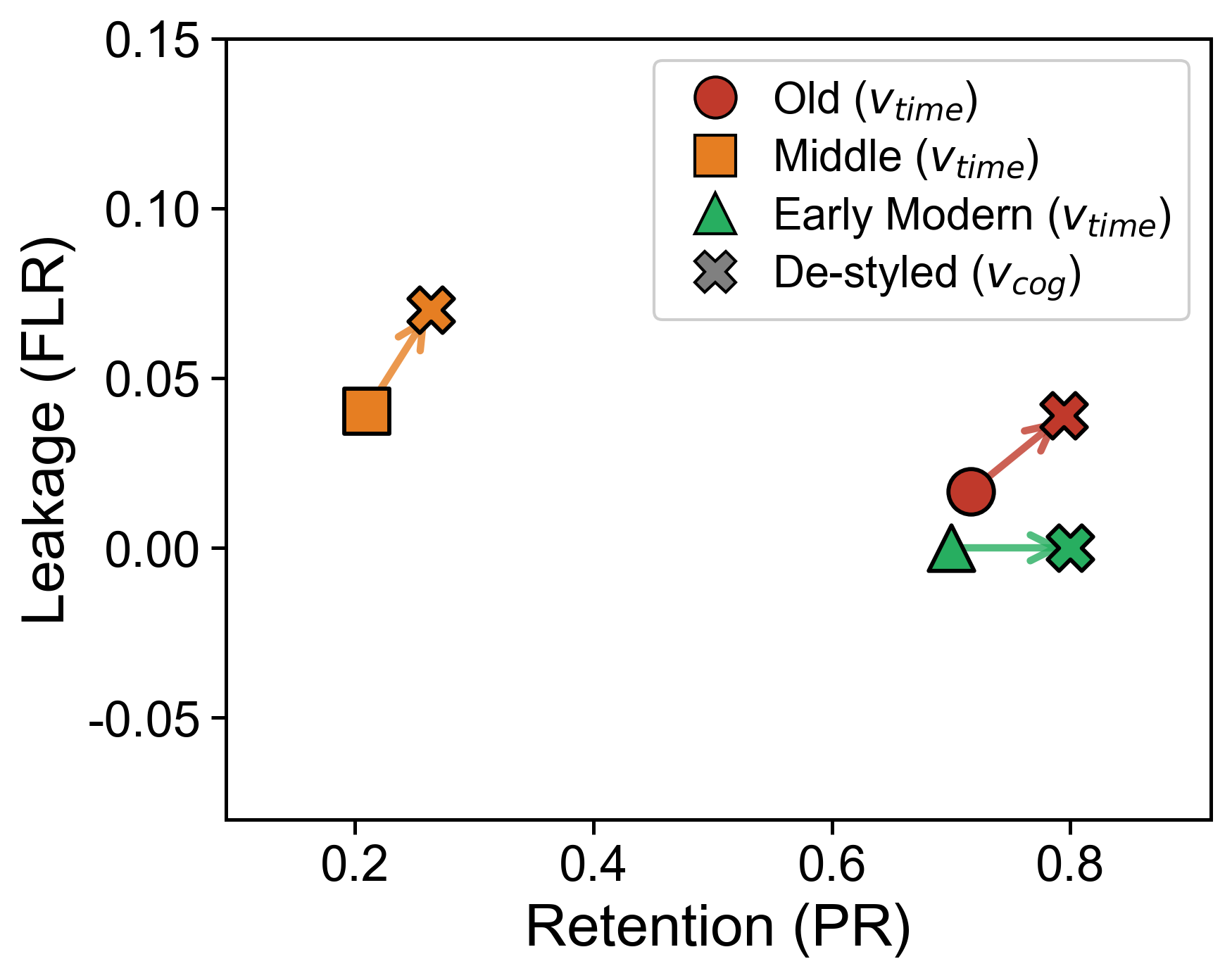}
        \centerline{\small (b) Vector Field for English}
    \end{minipage}
    \caption{Visualization of the entanglement vector fields. The arrows depict the shift in performance metrics when orthogonalizing the time vector to remove stylistic features ($\mathbf{v}_{time} \to \mathbf{v}_{cog}$).}
    \label{fig:entanglement_field}
\end{figure}

We evaluate the independence of temporal features using the disentanglement mechanism in Sec.~\ref{ssec:method_disentangle}. Figure~\ref{fig:entanglement_field} presents the entanglement vector field, visualizing the interplay between stylistic precision and future leakage during cognitive isolation.

Results reveal a distinct, era-dependent chronotope entanglement. While aggregate metrics (Appendix~\ref{ssec:app_disentangle}) suggest that disentanglement improves performance across the timeline, a granular analysis exposes a specific divergence in the \textit{Old} era. Here, removing dominant stylistic components notably increases FLR, suggesting that for deep historical horizons, the representation of ``when'' is intrinsically bound to ``how.'' Neutralizing the linguistic register weakens the enforcement of chronological constraints. Conversely, for \textit{Middle} and \textit{Early Modern} eras, the trajectory stabilizes and even improves, driving the overall robustness observed in the EnsCMP mechanism.

\subsection{Cross-Lingual Temporal Transfer}
\label{sec:results_cross_lingual}

Finally, we investigate the universality of these diachronic representations. We postulate that if the geometric encoding of time is universal, the chronological manifold of one language should possess a structural alignment with that of another. To verify this, projection experiments are performed where the internal state of the model is steered using foreign historical signals (e.g., English vectors modulating Chinese prompts). Qualitative illustrations of such cross-lingual register modulation are presented in Table~\ref{chinese_verse} and Table~\ref{english_verse}.

Analyses in Sec.~\ref{sec:cross_lingual_analysis} and Appendix~\ref{ssec:app_crosslingual} uncover two governing mechanisms. Qualitatively, we observe representation override: temporal signals suppress prompt-induced priming to steer generation across languages, validating a robust, language-agnostic ``chronotope.'' Quantitatively, cross-architecture benchmarks show that while discrete alignment is brittle—prone to catastrophic forgetting—continuous manifold projections preserve semantic structure. This topological isomorphism enables zero-shot temporal transfer where discrete matching proves inadequate.

\section{Conclusion}

This work introduces the TTE, a framework demonstrating that chronological time in LLMs functions as a shared, continuous manifold rather than fragmented metadata. By uncovering a topological isomorphism between Chinese and English temporal subspaces, we establish that LLMs organize history through a universal geometric logic transcending linguistic boundaries. Our methodology unifies discrete steering with continuous trajectory modeling, enabling precise historical navigation and zero-shot transfer of temporal cognition. These findings reveal a ``chronotope entanglement,'' where linguistic registers serve as structural anchors for deep epistemic constraints. Ultimately, our evidence suggests that LLMs construct a coherent internal world model governed by a structured temporal dimension. By bridging historical linguistics with geometric interpretability, we provide a new paradigm for decoding the evolution of human discourse within machine intelligence.

\FloatBarrier

\section{Limitations}

We identify three constraints regarding the scope, precision, and granularity of the TTE framework:

\begin{itemize} 
    \item \textbf{Latent Density Dependence.} The efficacy of the engine is strictly bounded by the pre-trained latent territory of the model. While our framework successfully \textit{navigates} existing historical representations, it cannot \textit{synthesize} linguistic competence absent from the training data. Consequently, resource-scarce periods may suffer from extraction noise, yielding vectors that capture superficial lexical markers rather than deep grammatical structures.

    \item \textbf{Chronotope Entanglement.} While we demonstrate a ``Shared'' manifold, complete disentanglement between stylistic surface forms and deep cognitive constraints remains elusive. Our findings indicate that for architectures of the current generation, the representation of ``when'' is inextricably bound to ``how.'' This limits the feasibility of purely semantic temporal steering, as neutralizing the archaic style often degrades the enforcement of epistemic boundaries.

    \item \textbf{Manifold Approximation Granularity.} Our framework constructs a continuous spline trajectory to enable fluid navigation. This geometric modeling inherently prioritizes global coherence over local granularity, serving as a low-dimensional approximation of historical evolution. Consequently, the trajectory may smooth over high-frequency, non-linear discontinuities (such as abrupt linguistic revolutions), capturing the prevailing zeitgeist rather than granular, event-level shifts.
\end{itemize}


\bibliography{custom}

\newpage
\appendix

\section{Appendix}

\subsection{Performance Scaling across Architectures and Steering Mechanisms}
\label{app:additional_results_models_methods}

In this appendix, we expand our analysis beyond the specific case study of Qwen2.5-14B-Base model presented in the main text. By benchmarking four distinct architectures (Qwen2.5-14B-Base (Qwen-14B), Qwen2.5-14B-Instruct (Qwen-14B-Inst), Llama-3.1-8B-Instruct (Llama-8B-Inst), and Gemma-2-9B-Instruct (Gemma-9B-Inst) across three interpretability dimensions—World Knowledge, Style-Cognition Disentanglement, and Cross-Lingual Transfer—we uncover the fundamental geometric laws governing temporal representations in LLMs.

In Tables \ref{tab:epistemic_flr}--\ref{tab:ghost_pr}, all quantitative results are reported as mean $\pm$ standard deviation, aggregated across three historical eras (\textit{Old, Middle, Early Modern}). The reported values represent the arithmetic mean of the era-specific scores and standard deviations, respectively, quantifying performance variability across evaluation samples. We benchmark the four steering mechanisms defined in Sec.~\ref{ssec:manifold_construction}: CAA (Contrastive Activation Addition), Ens (Ensemble CAA), CMP (Chronological Manifold Projection), and EnsCMP (Ensemble Manifold Projection).

\subsubsection{Principle I: The Geometry of Epistemic Boundaries}
\label{ssec:app_epistemic}

We first examine whether the geometric smoothing of temporal signals enhances the ability of the model to discern historical facts. Tables \ref{tab:epistemic_flr} and \ref{tab:epistemic_pr} present the Future Leakage Rate (FLR) and Precision Rate (PR).

The data reveals a distinct alignment-geometry trade-off. For the unaligned Qwen-14B base model, the manifold-based strategy (EnsCMP) significantly outperforms the discrete CAA baseline. Specifically, EnsCMP reduces the FLR to $0.260 \pm 0.319$ compared to $0.338 \pm 0.357$ for CAA, while simultaneously boosting the PR from $0.362$ to $0.607$. This suggests that pre-trained latent spaces contain stochastic noise; the manifold fitting process acts as a geometric filter, smoothing out irregularities to reconstruct a coherent temporal progression.

Conversely, instruction-tuned models exhibit a diminished reliance on geometric smoothing. For Qwen-14B-Inst, the discrete ensemble method (Ens) achieves the optimal balance (FLR $0.299$, PR $0.684$), indicating that alignment techniques may cluster temporal concepts, rendering complex spline fitting less critical. Furthermore, Llama-8B-Inst and Gemma-9B-Inst show a preference for the raw CAA baseline, with Llama achieving its lowest FLR ($0.187$) via discrete modulation. This implies that strong instruction tuning linearizes the latent space, potentially making the "Time" feature sufficiently discrete such that manifold projection offers diminishing returns or introduces unnecessary interpolation noise.

\begin{table*}[ht]
\centering

\setlength{\tabcolsep}{5pt}
\begin{tabular}{lcccc}
\toprule
\textbf{Model} & \textbf{CAA} & \textbf{Ens} & \textbf{CMP} & \textbf{EnsCMP} \\
\midrule
Qwen-14B & $0.338 \pm 0.357$ & $0.428 \pm 0.314$ & $0.324 \pm 0.384$ & $\mathbf{0.260 \pm 0.319}$ \\
Qwen-14B-Inst & $0.370 \pm 0.279$ & $\mathbf{0.299 \pm 0.211}$ & $0.364 \pm 0.246$ & $0.356 \pm 0.291$ \\
Llama-8B-Inst & $\mathbf{0.187 \pm 0.237}$ & $0.307 \pm 0.289$ & $0.316 \pm 0.435$ & $0.420 \pm 0.386$ \\
Gemma-9B-Inst & $\mathbf{0.260 \pm 0.201}$ & $0.331 \pm 0.228$ & $0.329 \pm 0.246$ & $0.449 \pm 0.320$ \\
\bottomrule
\end{tabular}
\caption{World knowledge integrity analysis. Future Leakage Rate (FLR) measures the fraction of generated entities that belong to the future relative to the target era. Lower values indicate better suppression of anachronisms.}
\label{tab:epistemic_flr}
\end{table*}

\begin{table*}[ht]
\centering

\setlength{\tabcolsep}{5pt}
\begin{tabular}{lcccc}
\toprule
\textbf{Model} & \textbf{CAA} & \textbf{Ens} & \textbf{CMP} & \textbf{EnsCMP} \\
\midrule
Qwen-14B & $0.362 \pm 0.367$ & $0.505 \pm 0.352$ & $0.543 \pm 0.411$ & $\mathbf{0.607 \pm 0.354}$ \\
Qwen-14B-Inst & $0.613 \pm 0.274$ & $\mathbf{0.684 \pm 0.264}$ & $0.535 \pm 0.296$ & $0.644 \pm 0.291$ \\
Llama-8B-Inst & $\mathbf{0.597 \pm 0.352}$ & $0.559 \pm 0.232$ & $0.567 \pm 0.462$ & $0.530 \pm 0.371$ \\
Gemma-9B-Inst & $\mathbf{0.740 \pm 0.201}$ & $0.652 \pm 0.235$ & $0.621 \pm 0.217$ & $0.518 \pm 0.286$ \\
\bottomrule
\end{tabular}
\caption{World knowledge integrity analysis. Precision Rate (PR) measures the fraction of generated entities that are historically accurate (in-scope) for the target era. Higher values indicate richer recall of era-appropriate entities.}
\label{tab:epistemic_pr}
\end{table*}

\subsubsection{Principle II: The Robustness of the Chronotope}
\label{ssec:app_disentangle}

To verify whether the steering relies on superficial style or deep cognition, we compare the performance of the full time vector ($v_{time}$) against the orthogonalized cognitive vector ($v_{cog}$). 

Tables \ref{tab:disentangle_flr_time}--\ref{tab:disentangle_pr_cog} reveal the superior cognitive persistence of manifold-based methods. For Qwen-14B, when moving from $v_{time}$ to $v_{cog}$ (removing style), the EnsCMP method not only maintains an extremely low FLR ($0.193 \to 0.169$) but actually improves its PR ($0.707 \to 0.747$). In contrast, discrete methods like CAA exhibit stagnation or degradation ($\text{FLR} \approx 0.30$) when style is stripped. This confirms that the manifold projection method captures a stable "chronotope" that enforces epistemic boundaries even without stylistic cues.

This robustness extends across architectures with varying degrees of entanglement. For Llama-8B-Inst, EnsCMP demonstrates remarkable stability, maintaining a virtually constant PR ($0.665 \to 0.668$) during disentanglement, whereas discrete methods show greater fluctuation. Conversely, Gemma-9B-Inst exhibits stronger intrinsic entanglement, where removing style generally triggers an increase in future leakage across all methods; however, manifold strategies (CMP) mitigate this degradation significantly better than ensemble approaches (e.g., CMP FLR shifts $0.221 \to 0.242$ vs. Ens $0.211 \to 0.312$). Overall, manifold-based steering consistently offers the most reliable decoupling of historical cognition from surface form across diverse model families.

\begin{table*}[t]
\centering

\setlength{\tabcolsep}{5pt}
\begin{tabular}{lcccc}
\toprule
\textbf{Model} & \textbf{CAA} & \textbf{Ens} & \textbf{CMP} & \textbf{EnsCMP} \\
\midrule
Qwen-14B & $0.290 \pm 0.262$ & $0.390 \pm 0.244$ & $0.238 \pm 0.273$ & $\mathbf{0.193 \pm 0.199}$ \\
Qwen-14B-Inst & $0.335 \pm 0.266$ & $\mathbf{0.264 \pm 0.137}$ & $0.385 \pm 0.307$ & $0.384 \pm 0.242$ \\
Llama-8B-Inst & $\mathbf{0.140 \pm 0.153}$ & $0.285 \pm 0.200$ & $0.244 \pm 0.216$ & $0.193 \pm 0.183$ \\
Gemma-9B-Inst & $\mathbf{0.186 \pm 0.116}$ & $0.211 \pm 0.172$ & $0.221 \pm 0.107$ & $0.276 \pm 0.225$ \\
\bottomrule
\end{tabular}
\caption{Disentanglement baseline using the full time vector ($\mathbf{v}_{time}$). FLR evaluates the performance of this raw vector, which retains both stylistic and cognitive components. Lower values indicate better suppression.}
\label{tab:disentangle_flr_time}
\end{table*}

\begin{table*}[t]
\centering

\setlength{\tabcolsep}{5pt}
\begin{tabular}{lcccc}
\toprule
\textbf{Model} & \textbf{CAA} & \textbf{Ens} & \textbf{CMP} & \textbf{EnsCMP} \\
\midrule
Qwen-14B & $0.299 \pm 0.323$ & $0.381 \pm 0.268$ & $0.208 \pm 0.229$ & $\mathbf{0.169 \pm 0.191}$ \\
Qwen-14B-Inst & $0.324 \pm 0.225$ & $\mathbf{0.279 \pm 0.139}$ & $0.417 \pm 0.290$ & $0.345 \pm 0.278$ \\
Llama-8B-Inst & $\mathbf{0.156 \pm 0.135}$ & $0.290 \pm 0.244$ & $0.321 \pm 0.231$ & $0.216 \pm 0.202$ \\
Gemma-9B-Inst & $0.245 \pm 0.154$ & $0.312 \pm 0.176$ & $\mathbf{0.242 \pm 0.090}$ & $0.359 \pm 0.219$ \\
\bottomrule
\end{tabular}
\caption{Disentanglement efficacy using the cognitive vector ($\mathbf{v}_{cog}$). FLR evaluates performance after stylistic features have been removed to isolate cognitive content. Lower values indicate robust cognitive persistence.}
\label{tab:disentangle_flr_cog}
\end{table*}

\begin{table*}[!htbp]
\centering

\setlength{\tabcolsep}{5pt}
\begin{tabular}{lcccc}
\toprule
\textbf{Model} & \textbf{CAA} & \textbf{Ens} & \textbf{CMP} & \textbf{EnsCMP} \\
\midrule
Qwen-14B & $0.676 \pm 0.290$ & $0.568 \pm 0.288$ & $0.620 \pm 0.295$ & $\mathbf{0.707 \pm 0.261}$ \\
Qwen-14B-Inst & $0.607 \pm 0.261$ & $\mathbf{0.736 \pm 0.137}$ & $0.473 \pm 0.284$ & $0.599 \pm 0.258$ \\
Llama-8B-Inst & $\mathbf{0.760 \pm 0.224}$ & $0.682 \pm 0.172$ & $0.748 \pm 0.217$ & $0.665 \pm 0.197$ \\
Gemma-9B-Inst & $\mathbf{0.814 \pm 0.116}$ & $0.789 \pm 0.172$ & $0.771 \pm 0.130$ & $0.691 \pm 0.214$ \\
\bottomrule
\end{tabular}
\caption{Disentanglement baseline using the full time vector ($\mathbf{v}_{time}$). PR reflects the accurate retrieval of historical facts using this raw vector. Higher values indicate richer recall.}
\label{tab:disentangle_pr_time}
\end{table*}

\begin{table*}[!htbp]
\centering

\setlength{\tabcolsep}{5pt}
\begin{tabular}{lcccc}
\toprule
\textbf{Model} & \textbf{CAA} & \textbf{Ens} & \textbf{CMP} & \textbf{EnsCMP} \\
\midrule
Qwen-14B & $0.659 \pm 0.334$ & $0.577 \pm 0.305$ & $0.650 \pm 0.312$ & $\mathbf{0.747 \pm 0.263}$ \\
Qwen-14B-Inst & $0.651 \pm 0.230$ & $\mathbf{0.721 \pm 0.139}$ & $0.558 \pm 0.307$ & $0.580 \pm 0.312$ \\
Llama-8B-Inst & $\mathbf{0.744 \pm 0.215}$ & $0.676 \pm 0.199$ & $0.671 \pm 0.227$ & $0.668 \pm 0.234$ \\
Gemma-9B-Inst & $\mathbf{0.755 \pm 0.154}$ & $0.688 \pm 0.176$ & $0.758 \pm 0.090$ & $0.640 \pm 0.219$ \\
\bottomrule
\end{tabular}
\caption{Disentanglement efficacy using the cognitive vector ($\mathbf{v}_{cog}$). PR measures the retention of historical facts after stylistic features have been stripped. Higher values indicate minimal knowledge loss.}
\label{tab:disentangle_pr_cog}
\end{table*}

\subsubsection{Principle III: Topology is the Universal Translator}
\label{ssec:app_crosslingual}

We further scrutinize universality by separating direct modulation (transferring raw source language vectors to target prompts without alignment) and manifold-aligned modulation (transferring linearly aligned source language vectors via procrustes rotation to target prompts).

The contrast between FLR (Tables \ref{tab:direct_flr}, \ref{tab:ghost_flr}) and PR (Tables \ref{tab:direct_pr}, \ref{tab:ghost_pr}) provides the strongest evidence for the topological isomorphism of the manifold. 

For Qwen-14B, the discrete CAA method fails catastrophically in the challenging manifold-aligned setting: its FLR spikes to $0.701$, and its PR collapses to $0.146$, indicating a total functional failure where the model neither suppresses future knowledge nor retains historical facts. In sharp contrast, EnsCMP maintains a robust FLR ($0.196$) and a healthy PR ($0.371$), demonstrating that the geometric curvature of the temporal manifold is shared across languages.

Crucially, this topological advantage extends to instruction-tuned models, albeit with different dynamics. For Llama-8B-Inst, while CAA achieves a low FLR in manifold-aligned modulation ($0.150$), it suffers from a significant degradation in PR ($0.320$), suggesting that discrete vectors may be suppressing capability indiscriminately. EnsCMP, however, recovers the PR to $0.512$ while maintaining effective suppression (FLR $0.204$). Similarly, for Qwen-14B-Inst, the Ensemble and EnsCMP methods consistently outperform CAA in PR across both direct and manifold-aligned settings. This pattern confirms that while discrete alignment (CAA) is brittle and prone to "catastrophic forgetting" during cross-lingual transfer, the continuous manifold strategies (EnsCMP) successfully preserve the semantic structure of time, balancing epistemic boundary enforcement with knowledge retention across diverse architectures.

\begin{table*}[ht]
\centering

\setlength{\tabcolsep}{5pt}
\begin{tabular}{lcccc}
\toprule
\textbf{Model} & \textbf{CAA} & \textbf{Ens} & \textbf{CMP} & \textbf{EnsCMP} \\
\midrule
Qwen-14B & $0.450 \pm 0.360$ & $0.459 \pm 0.389$ & $0.219 \pm 0.266$ & $\mathbf{0.168 \pm 0.226}$ \\
Qwen-14B-Inst & $0.379 \pm 0.359$ & $\mathbf{0.213 \pm 0.259}$ & $0.280 \pm 0.286$ & $0.304 \pm 0.297$ \\
Llama-8B-Inst & $\mathbf{0.111 \pm 0.204}$ & $0.198 \pm 0.288$ & $0.139 \pm 0.257$ & $0.124 \pm 0.186$ \\
Gemma-9B-Inst & $0.411 \pm 0.384$ & $\mathbf{0.381 \pm 0.399}$ & $0.389 \pm 0.381$ & $0.454 \pm 0.432$ \\
\bottomrule
\end{tabular}
\caption{Cross-lingual transfer using direct modulation. FLR is evaluated when transferring raw source language vectors directly to target prompts (e.g., English Vector $\to$ English Prompt).}
\label{tab:direct_flr}
\end{table*}

\begin{table*}[!hbtp]
\centering

\setlength{\tabcolsep}{5pt}
\begin{tabular}{lcccc}
\toprule
\textbf{Model} & \textbf{CAA} & \textbf{Ens} & \textbf{CMP} & \textbf{EnsCMP} \\
\midrule
Qwen-14B & $0.701 \pm 0.357$ & $0.473 \pm 0.364$ & $0.442 \pm 0.310$ & $\mathbf{0.196 \pm 0.213}$ \\
Qwen-14B-Inst & $0.386 \pm 0.378$ & $\mathbf{0.240 \pm 0.271}$ & $0.321 \pm 0.283$ & $0.312 \pm 0.261$ \\
Llama-8B-Inst & $\mathbf{0.150 \pm 0.280}$ & $0.157 \pm 0.288$ & $0.341 \pm 0.405$ & $0.204 \pm 0.258$ \\
Gemma-9B-Inst & $\mathbf{0.474 \pm 0.401}$ & $0.610 \pm 0.445$ & $0.536 \pm 0.433$ & $0.542 \pm 0.412$ \\
\bottomrule
\end{tabular}
\caption{Cross-lingual transfer using manifold-aligned modulation. FLR is evaluated when transferring vectors across languages via Procrustes alignment (e.g., English Vector $\to$ Chinese Prompt).}
\label{tab:ghost_flr}
\end{table*}

\begin{table*}[!hbtp]
\centering

\setlength{\tabcolsep}{5pt}
\begin{tabular}{lcccc}
\toprule
\textbf{Model} & \textbf{CAA} & \textbf{Ens} & \textbf{CMP} & \textbf{EnsCMP} \\
\midrule
Qwen-14B & $0.310 \pm 0.338$ & $0.354 \pm 0.353$ & $0.174 \pm 0.248$ & $\mathbf{0.275 \pm 0.271}$ \\
Qwen-14B-Inst & $0.271 \pm 0.365$ & $\mathbf{0.527 \pm 0.359}$ & $0.327 \pm 0.381$ & $0.386 \pm 0.295$ \\
Llama-8B-Inst & $0.492 \pm 0.413$ & $0.468 \pm 0.405$ & $\mathbf{0.657 \pm 0.403}$ & $0.583 \pm 0.382$ \\
Gemma-9B-Inst & $\mathbf{0.396 \pm 0.381}$ & $0.349 \pm 0.380$ & $0.286 \pm 0.365$ & $0.227 \pm 0.331$ \\
\bottomrule
\end{tabular}
\caption{Cross-lingual transfer using direct modulation. PR is evaluated when using native language vectors without geometric rotation.}
\label{tab:direct_pr}
\end{table*}

\begin{table*}[!hbtp]
\centering

\setlength{\tabcolsep}{5pt}
\begin{tabular}{lcccc}
\toprule
\textbf{Model} & \textbf{CAA} & \textbf{Ens} & \textbf{CMP} & \textbf{EnsCMP} \\
\midrule
Qwen-14B & $0.146 \pm 0.228$ & $0.294 \pm 0.325$ & $0.189 \pm 0.237$ & $\mathbf{0.371 \pm 0.251}$ \\
Qwen-14B-Inst & $0.348 \pm 0.387$ & $\mathbf{0.510 \pm 0.345}$ & $0.356 \pm 0.363$ & $0.431 \pm 0.329$ \\
Llama-8B-Inst & $0.320 \pm 0.400$ & $0.446 \pm 0.411$ & $0.366 \pm 0.370$ & $\mathbf{0.512 \pm 0.407}$ \\
Gemma-9B-Inst & $\mathbf{0.289 \pm 0.336}$ & $0.130 \pm 0.270$ & $0.214 \pm 0.330$ & $0.178 \pm 0.304$ \\
\bottomrule
\end{tabular}
\caption{Cross-lingual transfer using manifold-aligned modulation. PR is evaluated when using foreign language vectors aligned via Procrustes analysis, testing the universality of the manifold.}
\label{tab:ghost_pr}
\end{table*}


\subsection{Cross-linguistic Generalization}
\label{sec:cross_lingual_analysis}

In this section, we present a detailed qualitative evaluation of the cross-linguistic experiments introduced in the main text. To demonstrate the generalization of our framework across genres, we here provide verse examples (complementing the prose instances in Sec.~\ref{ssec:results_manifold}) and examine how the projection of hidden states from input prompts onto historical subspaces constructed from disparate source languages affects the stylistic output.

Detailed comparisons are provided in Table~\ref{chinese_verse} and Table~\ref{english_verse}. To systematically analyze these effects, we define two distinct modulation settings:

\begin{itemize}
    \item \textbf{Native Signal Modulation}: The source of the steering vector shares the linguistic identity of the prompt. For example, this involves the modulation of a \textit{Modern Chinese} prompt using a vector extracted from \textit{Old Chinese} data (denoted as \textsc{ZH}$\rightarrow$\textsc{ZH}). This setting establishes a baseline for the capability of the model to perform monolingual diachronic style transfer.
    \item \textbf{Cross-Lingual Signal Modulation}: The source of the steering vector differs from the linguistic identity of the prompt. For instance, this involves the modulation of a \textit{Modern Chinese} prompt using a vector extracted from \textit{Old English} data (denoted as \textsc{EN}$\rightarrow$\textsc{ZH}). As detailed in Sec.~\ref{ssec:cross-lingual}, we evaluate this via direct modulation (raw vector integration) and manifold-aligned modulation (Procrustes-based alignment), rigorously testing the robustness of the time vector and the disentanglement between temporal features and linguistic surface forms.
\end{itemize}

As evidenced in Table~\ref{chinese_verse}, we observe a compelling phenomenon of representation override: the integration of the \textit{Old English} steering vector into the processing of \textit{Modern Chinese} prompts results in outputs generated in the \textit{Old English} register. Here, the steering magnitude of the time vector effectively suppresses the linguistic priming provided by the Chinese prompt, shifting the generation trajectory entirely across language boundaries. 

Conversely, as detailed in Table~\ref{english_verse}, modulating \textit{Modern English} contexts with \textit{Old Chinese} signals induces outputs in \textit{Classical Chinese}. These results suggest that the extracted time vectors encode a robust ``chronotope''—a unified representation of time and language—that is sufficiently potent to dictate the generation path, demonstrating that the temporal geometry within the model acts as an active, traversable dimension capable of zero-shot stylistic transfer.


\begin{table*}[ht]
    \centering

    \renewcommand{\arraystretch}{1.15}
    \begin{tabularx}{\linewidth}{|X|X|X|}
        \hline

        \rowcolor{promptbg}
        \multicolumn{3}{|l|}{\textbf{PROMPT}} \\
        \hline
        \rowcolor{promptbg}
        \multicolumn{3}{|l|}{请以诗歌的形式描述一场冬日的暴风雪。} \\
        \rowcolor{promptbg}
        \multicolumn{3}{|l|}{\trans{Please describe a winter storm in the form of a poem.}} \\
        \hline

        \textbf{Baseline (\textit{Modern State})} &
        \textbf{Native Signal (ZH$\rightarrow$ZH)} &
        \textbf{Cross-Lingual Signal (EN$\rightarrow$ZH)} \\
        \hline

        \rowcolor{injectionbg}
        \multicolumn{3}{|l|}{\textbf{Signal: \textit{Old}}} \\
        \hline
        冬夜寒风起，雪花漫天飞。银装素裹中，世界静悄悄... \par
        \trans{Winter winds rise in the night, snowflakes fly across the sky. Amidst the silver-clad world, all is silent...} &
        朔风怒号兮，天地昏黄，素羽纷飞兮，六出其芒... \par
        \trans{The north wind howls—lo! Heaven and earth turn dusky yellow. White plumes scatter—lo! Six-pointed are their blades...} &
        þurfe geong wundorlic þing, wolcenum geweorþeð on wolde, hrerende hræfena gehyngum hrimbealo sceawade. \par
        \trans{A wondrous thing comes to pass in the clouds upon the wold; stirring the ravens' hunger, the frost-ruin appears.} \\
        \hline

        \rowcolor{injectionbg}
        \multicolumn{3}{|l|}{\textbf{Signal: \textit{Middle}}} \\
        \hline
        冬夜寒风起，雪花漫天飞。银装素裹中，世界静悄悄... \par
        \trans{Winter winds rise in the night...} &
        朔风卷地雪漫天，万木凋零玉絮旋。冻合江河冰作岸，云迷岩谷霰连绵... \par
        \trans{The north wind rolls across the earth, snow fills the sky; Ten thousand trees wither, jade fluffs swirl on high. Ice locks the rivers, forming banks of frost; Clouds confuse the valleys, in endless sleet lost.} &
        The wynter storme is so wode, The snow floures ful thikke falleth, As hevy stones doun they shalle... \par
        \trans{The winter storm is so furious; The snow flowers fall full thick, As heavy stones down they shall [fall]...} \\
        \hline

        \rowcolor{injectionbg}
        \multicolumn{3}{|l|}{\textbf{Signal: \textit{Early Modern}}} \\
        \hline
        冬夜寒风起，雪花漫天飞。银装素裹中，世界静悄悄... \par
        \trans{Winter winds rise in the night...} &
        纷纷扬扬漫天，素裹银装玉栏，风卷残云舞翩跹... \par
        \trans{Fluttering and scattering across the vast sky, Clad in white, silver-wrapped upon jade balustrades; The wind coils the scattered clouds, dancing in a whirl...} &
         Thou art beholding unto Winter's fierce tempest, As ſnow doth ſwiftly ſwell from heaven above... \par
       \trans{You are beholding Winter's fierce tempest, As snow does swiftly swell from heaven above...} \\
        \hline

    \end{tabularx}
    \caption{Qualitative demonstration of diachronic register modulation via TTE. Baseline is compared against native-signal and cross-lingual-signal modulation. Text in gray represents English translations.}
    \label{chinese_verse}
\end{table*}

\begin{table*}[ht]
    \centering

    \renewcommand{\arraystretch}{1.15}
    \begin{tabularx}{\linewidth}{|X|X|X|}
        \hline

        \rowcolor{promptbg}
        \multicolumn{3}{|l|}{\textbf{PROMPT}} \\
        \hline
        \rowcolor{promptbg}
        \multicolumn{3}{|l|}{Describe a winter storm in verse.} \\
        \hline

        \textbf{Baseline (\textit{Modern State})} &
        \textbf{Native Signal (EN$\rightarrow$EN)} &
        \textbf{Cross-Lingual Signal (ZH$\rightarrow$EN)} \\
        \hline

        \rowcolor{injectionbg}
        \multicolumn{3}{|l|}{\textbf{Signal: \textit{Old}}} \\
        \hline
        The sky was dark and heavy, A blanket of grey, so dreary... &
        三月春寒，雪落纷纷，北风呼啸，天地昏昏...\par
        \trans{Spring chill in March, snow falls in profusion; The north wind whistles, heaven and earth are dim...} &
        惟天有雪，其寒若霜。风兮何起，云兮何茫... \par
        \trans{Heaven alone holds the snow; its cold is like frost. Oh, from where does the wind rise? Oh, how vast are the clouds...} \\
        \hline

        \rowcolor{injectionbg}
        \multicolumn{3}{|l|}{\textbf{Signal: \textit{Middle}}} \\
        \hline
         The sky was dark and heavy, A blanket of grey, so dreary... &
        þe wynd doth blowe wyth fyer and might, þe snow falleþ downe styffe and bryght... \par
        \trans{The wind blows with fire and might, The snow falls down stiff and bright...} &
        惟馀雪皓然，白者皆素焉。寒月霜气冽，清风松韵传... \par
        \trans{Only the snow remains bright; all that is white is pure. The cold moon's frost-air is biting; the clear wind conveys the pine's rhyme...} \\
        \hline

        \rowcolor{injectionbg}
        \multicolumn{3}{|l|}{\textbf{Signal: \textit{Early Modern}}} \\
        \hline
         The sky was dark and heavy, A blanket of grey, so dreary... &
        A ſtrong wiſterne winde doth blow, And with his ſtrength he ſendes forth ſnow... \par
        \trans{A strong western wind does blow, And with his strength he sends forth snow...} &
        南枝梅蕊，北窗竹影，西阁茶烟。雪风儿六花作团,云月儿一玉相联... \par
        \trans{Plum blossoms on the south branch, bamboo shadows by the north window, tea smoke in the west pavilion. The snowy wind swirls six-petaled flowers into balls; clouds and moon join in jade-like unity...} \\
        \hline

    \end{tabularx}
    \caption{Qualitative demonstration of diachronic register modulation via TTE. Baseline is compared against native-signal and cross-lingual-signal modulation.}
    \label{english_verse}
\end{table*}

\subsection{Model details}
\label{model_detail}

This appendix reports the architectural specifications of the four LLMs used in our experiments (Table~\ref{tab:model_specs}). We select the Qwen2.5-14B Base/Instruct pair specifically to isolate the impact of post-training alignment on temporal geometry. Furthermore, we extend our evaluation to Llama-3.1-8B-Instruct and Gemma-2-9B-Instruct to verify the cross-architectural universality of the chronological manifold beyond a single model family.

\begin{table*}[t]
\centering
\setlength{\tabcolsep}{4pt}
\renewcommand{\arraystretch}{1.12}
\begin{tabular}{lccccc}
\hline
\textbf{Model} & \textbf{Params} & \textbf{Layers} & \textbf{Attention Heads} & \textbf{Hidden States} & \textbf{Feedforward} \\
\hline
Gemma-2-9B-Instruct          & 9B  & 42 & 16 & 3584 & 14336 \\
Llama-3.1-8B-Instruct           & 8B  & 32 & 32 & 4096 & 14336 \\
Qwen2.5-14B-Instruct   & 14B & 48 & 40 & 5120 & 13824 \\
Qwen2.5-14B-Base         & 14B & 48 & 40 & 5120 & 13824 \\
\hline
\end{tabular}
\caption{Model specifications.}
\label{tab:model_specs}
\end{table*}

\subsection{Prompts for Temporal State Extraction}
\label{app:prompts}

This appendix documents the immersive tasks employed to extract temporal states via CAA. To ensure the cross-lingual consistency of the extracted signals, these tasks are designed with semantic parallelism across Chinese and English, as detailed in Table~\ref{tab:short_tasks_zh} and Table~\ref{tab:short_tasks_en}.


\begin{table*}[t]
\centering

\setlength{\tabcolsep}{4pt}
\renewcommand{\arraystretch}{1.15}

\begin{tabularx}{\textwidth}{
p{0.06\textwidth}
>{\raggedright\arraybackslash}X
}
\toprule
\textbf{\#} & \textbf{Task} \\
\midrule
1  & 用三句话介绍你自己、你的身份和你每日所见的制度。 \\
2  & 描写春日清晨城郭或乡野的景色。 \\
3  & 解释什么是'薄如纸而能映人像与远景，可收于衣囊'的怪物。 \\
4  & 说明如何度量时间、距离、重量，常用何种单位？ \\
5  & 城中多人发热咳嗽而亡，你如何判断病因并建议官府采取措施？ \\
6  & 连续歉收或大水成灾，你认为成因为何？如何赈济安民？ \\
7  & 描述一次远行的准备工作和沿途可能遇到的困难。 \\
8  & 解释天象变化（日食、彗星）的含义和应对方法。 \\
9  & 介绍当世最重要的三部典籍及其核心思想。 \\
10 & 描述一场婚礼或丧葬仪式的基本流程。 \\
11 & 说明如何治疗常见的风寒发热之症。 \\
12 & 解释为何某些行为会被视为违礼或犯法。 \\
13 & 描述一位德高望重之人应具备的品质。 \\
14 & 说明农事的基本节气安排和耕作要点。 \\
15 & 描写月夜独坐或登高远眺的感受（诗意表达）。 \\
\bottomrule
\end{tabularx}
\caption{Chinese immersive tasks.}
\label{tab:short_tasks_zh}
\end{table*}

\begin{table*}[t]
\centering

\setlength{\tabcolsep}{4pt}
\renewcommand{\arraystretch}{1.15}

\begin{tabularx}{\textwidth}{
p{0.06\textwidth}
>{\raggedright\arraybackslash}X
}
\toprule
\textbf{\#} & \textbf{Task} \\
\midrule
1  & Introduce yourself in three sentences: your identity and the institutions you encounter daily. \\
2  & Describe a spring morning scene near a town or countryside. \\
3  & Explain what a 'paper-thin object that reflects faces and distant scenes, fitting in a pocket' might be. \\
4  & Explain how you measure time, distance, and weight. What units are commonly used? \\
5  & A fever and coughing illness kills many in town. How do you determine the cause and advise authorities? \\
6  & Continuous crop failure or flooding occurs. What causes do you suspect and how to provide relief? \\
7  & Describe preparations for a long journey and difficulties you might encounter. \\
8  & Explain the meaning of celestial changes (eclipses, comets) and how to respond. \\
9  & Introduce the three most important texts of your era and their core ideas. \\
10 & Describe the basic procedures of a wedding or funeral ceremony. \\
11 & Explain how to treat a common cold or fever. \\
12 & Explain why certain behaviors are considered improper or illegal. \\
13 & Describe the qualities a person of great virtue should possess. \\
14 & Explain the basic agricultural calendar and farming essentials. \\
15 & Describe feelings while sitting alone on a moonlit night or gazing from a height (poetic expression). \\
\bottomrule
\end{tabularx}
\caption{English immersive tasks.}
\label{tab:short_tasks_en}
\end{table*}

\subsection{Contrastive Style Dataset}
\label{app:style_pairs}

To strictly distinguish between linguistic style and temporal cognition, we curated a contrastive style dataset. This dataset consists of parallel semantic pairs where the core meaning remains invariant while the linguistic register shifts across four canonical eras: \textit{Old}, \textit{Middle}, \textit{Early Modern}, and \textit{Modern}. 

These pairs serve as the basis for constructing the style subspace ($\mathbf{U}_{style}$) used in the disentanglement experiments (Sec.~\ref{ssec:method_disentangle}). We isolate the geometric direction of 'style' by applying PCA to the difference vectors between modern and archaic representations.

Table~\ref{tab:style_pairs_samples} presents representative examples from the dataset, spanning diverse topics such as time, nature, and social customs. 

\begin{table*}[t]
\centering

\setlength{\tabcolsep}{4pt}
\renewcommand{\arraystretch}{1.15}

\begin{tabularx}{\textwidth}{
    p{0.12\textwidth}
    >{\raggedright\arraybackslash}X
    >{\raggedright\arraybackslash}X
}
\toprule
\textbf{Period} & \textbf{Chinese Text} & \textbf{English Text} \\
\midrule

\multicolumn{3}{c}{\textit{Topic: Passage of Time}} \\
\midrule
\textit{Old} & 逝者如斯夫，不舍昼夜。 & Tīd ne bīdeþ nǣnigne mann, hēo æfre forðgæþ. \\
\textit{Middle} & 光阴似箭，日月如梭，岁不我与，时不再来。 & Tyme abideth noon, it passeth as a streem. \\
\textit{E. Mod} & 俗话说得好，一寸光阴一寸金，寸金难买寸光阴哪。 & Time and Tide wait for no Man, but ever flow onward. \\
\textit{Modern} & 时间不等人。 & Time waits for no one. \\
\midrule

\multicolumn{3}{c}{\textit{Topic: Blooming Flowers}} \\
\midrule
\textit{Old} & 百花竞放，其香四溢。 & Blostman springaþ, hira wlite is fæger. \\
\textit{Middle} & 繁花似锦，争奇斗艳，满园春色关不住。 & The floures bloomen faire, of dyverse colours and swete smell. \\
\textit{E. Mod} & 花儿都开了，红的白的紫的，好不热闘，香气扑鼻。 & The Flowers are blooming in great abundance, their sweet perfume fills the air. \\
\textit{Modern} & 花开了。 & The flowers are blooming. \\
\midrule

\multicolumn{3}{c}{\textit{Topic: Autumn Scenery (Verse)}} \\
\midrule
\textit{Old} & 蒹葭苍苍，白露为霜。秋风萧瑟，草木摇落。 & Hærfest is cumen, hēah wind blǣwþ; lēaf feallaþ, lond wearþ cald. \\
\textit{Middle} & 碧云天外雁南飞，黄叶纷纷满地堆。萧瑟西风吹不尽，一年秋色到楼台。 & The leves fallen from the tre, The wynde doth blowe so colde and kene; The somer dayes now fledde be, And al the worlde is bare and lene. \\
\textit{E. Mod} & 西风吹落叶，黄叶满庭阶。一年好景处，秋色最堪怀。 & When Autumn comes with golden Hue, The Leaves do fall from every Tree; The Summer Days have bid Adieu, And Nature mourns what used to be. \\
\textit{Modern} & 秋叶飘落，铺满了小径。风吹过树梢，带走了夏天。 & Autumn leaves are falling, carpeting the ground in gold. The wind whispers through bare branches, a farewell to warmth. \\
\midrule

\multicolumn{3}{c}{\textit{Topic: Wine and Gathering (Verse)}} \\
\midrule
\textit{Old} & 我有嘉宾，鼓瑟吹笙。呦呦鹿鸣，食野之苹。 & Wīn bringþ glædnesse, gāstas āhebbaþ; drync and drēam, dēorwyrðe sǣl. \\
\textit{Middle} & 葡萄美酒夜光杯，欲饮琵琶马上催。醉卧沙场君莫笑，古来征战几人回。 & Wyn bryngeth joye and maketh glad, It lyghteth up the derke of nyght; It comforteth the herte that's sad, And turneth wronge to alle aryght. \\
\textit{E. Mod} & 劝君更尽一杯酒，西出阳关无故人。今朝有酒今朝醉，明日愁来明日愁。 & Wine brings Joy to ev'ry Heart, It lifts the Soul and cheers the Mind; From Sorrow's Grip it sets apart, And leaves our Worldly Cares behind. \\
\textit{Modern} & 一杯美酒，带来欢乐。朋友相聚，笑语盈盈。 & A cup of wine brings joy. Friends gather, laughter fills the air. \\

\bottomrule
\end{tabularx}
\caption{Representative samples from the contrastive style dataset used for computing the style subspace. The dataset aligns semantic content across four historical periods.}
\label{tab:style_pairs_samples}
\end{table*}

\subsection{Evaluation Datasets Details}
\label{evaluation}

To comprehensively evaluate the effectiveness of the TTE framework, we constructed three specialized datasets, each targeting a specific dimension of temporal alignment: epistemic boundaries, causal reasoning, and feature disentanglement. The following sections provide representative samples and descriptions for each benchmark.

\subsubsection{Epistemic Cutoff Dataset}
This dataset is designed to probe the chronological knowledge boundaries of the model. It consists of open-ended queries requesting lists of entities (e.g., philosophers, inventions, tools). The ground truth for these queries is strictly segmented by historical era. For instance, when the model is steered to the \textit{Old} era (e.g., Pre-Qin), the generation of an entity from the \textit{Modern} era (e.g., "iPhone" or "Kant") is penalized as a future leakage (measured by FLR), while entities belonging to or preceding the target era are counted towards the PR.

\begin{table*}[h!]
\centering

\renewcommand{\arraystretch}{1.3}
\begin{tabular}{p{0.45\textwidth} p{0.45\textwidth}}
\hline
\textbf{English Query} & \textbf{Chinese Query} \\
\hline
Who are famous philosophers? & 著名的哲学家有哪些？ \\
What were major inventions? & 重大发明有哪些？ \\
What scripts were primarily used? & 主要书写系统有哪些？ \\
How was time kept? & 记录时间的主要工具是什么？ \\
How did people communicate over long distances? & 主要远程通讯方式是什么？ \\
\hline
\end{tabular}
\caption{Sample queries from the Epistemic Cutoff Dataset. The model is expected to filter its output based on the temporal knowledge cutoff of the steered era.}
\label{tab:epistemic_samples}
\end{table*}

\subsubsection{Causal Remodeling Dataset}
The Causal Remodeling Dataset evaluates whether the internal reasoning logic of the model shifts to align with the technological and causal constraints of the target era. Distinct from the Epistemic dataset which probes static factual retrieval, this dataset presents practical, daily-life scenarios (e.g., lighting, communication, preservation) that demand era-appropriate problem-solving. We apply the same FLR and PR metrics to this context: FLR measures the intrusion of anachronistic technologies (e.g., suggesting a ``flashlight'' in the Old era), while PR quantifies the successful adoption of historically valid causal mechanisms (e.g., ``oil lamps'').

\begin{table*}[h!]
\centering
\small
\renewcommand{\arraystretch}{1.3}
\begin{tabular}{p{0.45\textwidth} p{0.45\textwidth}}
\hline
\textbf{English Query} & \textbf{Chinese Query} \\
\hline
I need to send a message from Beijing to London. How long will it take? & 我需要从北京给伦敦发送信息，需要多长时间？ \\
The night is dark. What do I do? & 夜晚无灯，我该如何照亮？ \\
I want to preserve fish for the winter. What should I do? & 我想保存鱼肉过冬，该怎么做？ \\
In winter it's cold. How do I keep the house warm? & 冬天寒冷，我如何使家中温暖？ \\
How can I cross a river? & 要过河，应使用什么方式？ \\
\hline
\end{tabular}
\caption{Sample queries from the Causal Remodeling Dataset. These prompts test whether the model adopts the technological and causal constraints of the target era.}
\label{tab:causal_samples}
\end{table*}

\subsubsection{Mismatch Entanglement Dataset} This dataset serves as a rigorous stress test to evaluate the disentanglement between linguistic surface realization and latent cognitive representations. The prompts deliberately introduce chronological paradoxes by querying contemporary scientific or technological concepts while the model is modulated by steering vectors derived from \textit{Old}, \textit{Middle}, or \textit{Early Modern} periods. Such a configuration enables an empirical verification of whether the TTE manifold can achieve independent control over the epistemic grounding (the internal knowledge state) and the stylistic register (the linguistic output mode). By isolating these dimensions, we assess the robustness of the temporal steering mechanism against the explicit modern semantic constraints embedded within the prompts.

\begin{table*}[h!]
\centering
\small
\renewcommand{\arraystretch}{1.3}
\begin{tabular}{p{0.45\textwidth} p{0.45\textwidth}}
\hline
\textbf{English Query} & \textbf{Chinese Query} \\
\hline
Please answer using modern English prose, what are the main methods for treating diseases today? & 请用现代汉语白话文体回答，当今时代治疗疾病的主要方法是什么？ \\
Please answer using modern English prose, what media are currently used to store information? & 请用现代汉语白话文体回答，目前常用的信息存储介质是什么？ \\
Please answer using modern English prose, what sports are popular currently? & 请用现代汉语白话文体回答，目前流行的运动是什么？ \\
Please answer using modern English verse, what water supply systems exist in the modern era? & 请用现代汉语诗歌文体回答，现代供水系统有哪些？ \\
Please answer using modern English verse, what painting materials were used in the present day? & 请用现代汉语诗歌文体回答，现今绘画材料有哪些？ \\
\hline
\end{tabular}
\caption{Sample queries from the Mismatch Entanglement Dataset. These queries create deliberate conflicts between the requested content (\textit{Modern}) and the specified style or the active time vector.}
\label{tab:mismatch_samples}
\end{table*}

\end{CJK*}

\end{document}